\documentclass[journal]{IEEEtran}
\ifCLASSINFOpdf
\else
\fi
\usepackage{graphicx}
\usepackage[table]{xcolor}
\usepackage{tcolorbox}
\usepackage{hhline}
\usepackage{hyperref}
\usepackage[margin=0.8in]{geometry}
\usepackage{caption}
\usepackage{subcaption}

\usepackage{array}

\usepackage{multirow}

\usepackage{amsmath}

\usepackage{amsthm}
\usepackage{amssymb}
\usepackage{diagbox}

\usepackage{algorithm}
\usepackage{algpseudocode}

\usepackage{footmisc}

\usepackage{scalerel}
\usepackage{tikz}
\usetikzlibrary{svg.path}

\definecolor{orcidlogocol}{HTML}{A6CE39}
\tikzset{
  orcidlogo/.pic={
    \fill[orcidlogocol] svg{M256,128c0,70.7-57.3,128-128,128C57.3,256,0,198.7,0,128C0,57.3,57.3,0,128,0C198.7,0,256,57.3,256,128z};
    \fill[white] svg{M86.3,186.2H70.9V79.1h15.4v48.4V186.2z}
                 svg{M108.9,79.1h41.6c39.6,0,57,28.3,57,53.6c0,27.5-21.5,53.6-56.8,53.6h-41.8V79.1z M124.3,172.4h24.5c34.9,0,42.9-26.5,42.9-39.7c0-21.5-13.7-39.7-43.7-39.7h-23.7V172.4z}
                 svg{M88.7,56.8c0,5.5-4.5,10.1-10.1,10.1c-5.6,0-10.1-4.6-10.1-10.1c0-5.6,4.5-10.1,10.1-10.1C84.2,46.7,88.7,51.3,88.7,56.8z};
  }
}

\newcommand\orcidicon[1]{\href{https://orcid.org/#1}{\mbox{\scalerel*{
\begin{tikzpicture}[yscale=-1,transform shape]
\pic{orcidlogo};
\end{tikzpicture}
}{|}}}}

\usepackage{hyperref}

\begin{document}
%
\title{Learn Fast, Segment Well: Fast Object Segmentation Learning on the iCub Robot}
%
%
%

\author{Federico~Ceola\textsuperscript{\orcidicon{0000-0002-2356-0946}},~\IEEEmembership{Student~Member,~IEEE,}
        Elisa~Maiettini\textsuperscript{\orcidicon{0000-0002-0127-3014}},~
        Giulia~Pasquale\textsuperscript{\orcidicon{0000-0002-7221-3553}}, 
        Giacomo~Meanti\textsuperscript{\orcidicon{0000-0002-4633-2954}},~
        Lorenzo~Rosasco\textsuperscript{\orcidicon{0000-0003-3098-383X}},~
        and~Lorenzo~Natale\textsuperscript{\orcidicon{0000-0002-8777-5233}},~\IEEEmembership{Senior~Member,~IEEE}
\thanks{Federico Ceola, Elisa Maiettini, Giulia Pasquale and Lorenzo Natale are with Humanoid Sensing and Perception (HSP), Istituto Italiano di Tecnologia (IIT), Genoa, Italy (email: {\tt\footnotesize name.surname@iit.it}).}
\thanks{Federico Ceola, Giacomo Meanti (email: {\tt\footnotesize giacomo.meanti@edu.unige.it}) and Lorenzo Rosasco are with Laboratory for Computational and Statistical Learning (LCSL), with Machine Learning Genoa Center (MaLGa) and with Dipartimento di Informatica, Bioingegneria, Robotica e Ingegneria dei Sistemi (DIBRIS), University of Genoa, Genoa, Italy.}
\thanks{Lorenzo Rosasco is also with Istituto Italiano di Tecnologia (IIT) and with Center for Brains, Minds and Machines (CBMM), Massachusetts Institute of Technology (MIT), Cambridge, MA (email: {\tt\footnotesize lrosasco@mit.edu}).}
}

%
%

\markboth{IEEE TRANSACTIONS ON ROBOTICS}%
{}
%



\newcommand{\federico}[1]{{\color{blue}#1}}

\maketitle

\begin{abstract}
The visual system of a robot has different requirements depending on the application: it may require high accuracy or reliability, be constrained by limited resources or need fast adaptation to dynamically changing environments. In this work, we focus on the instance segmentation task and provide a comprehensive study of different techniques that allow adapting an object segmentation model in presence of novel objects or different domains.

We propose a pipeline for fast instance segmentation learning designed for robotic applications where data come in stream. It is based on an hybrid method leveraging on a pre-trained CNN for feature extraction and fast-to-train Kernel-based classifiers. We also propose a training protocol that allows to shorten the training time by performing feature extraction during the data acquisition. We benchmark the proposed pipeline on two robotics datasets and we deploy it on a real robot, i.e. the iCub humanoid. To this aim, we adapt our method to an incremental setting in which novel objects are learned on-line by the robot. 

The code to reproduce the experiments is publicly available on GitHub\footnote{\url{https://github.com/hsp-iit/online-detection}}.
\end{abstract}


%
\IEEEpeerreviewmaketitle


\section{INTRODUCTION}
\label{sec:introduction}
\IEEEPARstart{P}{erceiving} the environment is the first step for a robot to interact with it. Robots may be required to solve different tasks, as for instance grasping an object, interacting with a human or navigate in the environment avoiding obstacles. 

Different applications have different requirements for the robot vision system. For example, for an application in which a robot interacts with a predefined set of objects, fast learning is not the primary requirement. On the other hand, when a robot is operating in a dynamic environment (for instance a service robot operating in a hospital, a supermarket or a domestic environment), fast adaptation is fundamental.

The \textit{computer vision} literature is progressing at fast pace providing algorithms for object detection and segmentation that are remarkably powerful. These methods, however, are mostly based on deep neural networks and quite demanding in terms of training samples and optimization time. For this reason, they are badly suited for applications in robotics that require fast adaptation. Because the dominant trend in computer vision is to push performance as much as possible, comparably little effort is spent to propose methods that are designed to reduce training time. To fill this gap, in this work, we propose a comprehensive analysis in which we study various techniques for adaptation on a novel task. In particular, we consider approaches based on deep neural networks and on a combination of deep neural networks and Kernel methods, focusing on the trade-off between training time and accuracy.

We target the instance segmentation problem which consists in classifying every pixel of an image as belonging to an instance of a known object or to the background. In particular, we consider the scenario in which the robot encounters new objects during its operation and it is required to adapt its vision system so that it is able to segment them after a learning session that is as short as possible. We observe that this scenario offers opportunities to shorten the training time, for example if we are able to perform some of the training steps (i.e., feature extraction) already during data acquisition, and we propose a new method that is specifically optimized to reduce training time without compromising performance. 

Specifically, we propose an instance segmentation pipeline which extends and improves our previous work \cite{ceola2020segm}. In~\cite{ceola2020segm}, we proposed a fast learning method for instance segmentation of novel objects. One limitation of that method was to rely on a pre-trained region proposal network. In this work, we address this by making the region proposal learning on-line too. While this improves performance, it leads to a more complex and longer training pipeline if addressed na\"ively as it is done in~\cite{ceola2020rpn}. To this aim, we propose an approximated training protocol which can be separated in two steps: \textit{(i)} feature extraction and \textit{(ii)} fast and simultaneous training of the proposed approaches for region proposal, object detection and mask prediction. We show that this allows to further reduce the training time in the aforementioned robotic scenario.

In addition, we provide an extensive experimental analysis to investigate the training time/accuracy trade-off on two public datasets (i.e., YCB-Video~\cite{xiang2018posecnn} and HO-3D~\cite{hampali2020honnotate}). In particular, we show that our method is much more accurate than~\cite{ceola2020segm}, while requiring a comparable training time. Moreover, the proposed method allows to obtain accuracy similar to conventional fine-tuning approaches, while being trained much faster.

In summary, the contributions of this work are:
\begin{itemize}
    \item We propose a new pipeline and training protocol for instance based object segmentation, which is specifically designed for fast, on-line training.
    \item We benchmark the obtained results on two robotics datasets, namely YCB-Video~\cite{xiang2018posecnn} and HO-3D~\cite{hampali2020honnotate}.
    \item We provide an extensive study to compare our pipeline against conventional fine-tuning techniques, with an in-depth analysis of the trade-off between the required training time and the achieved accuracy.
    \item We deploy and demonstrate the proposed training pipeline on the iCub~\cite{icub} humanoid robot, adapting the algorithm for an incremental setting where target classes are not known a-priori.
\end{itemize}

This paper is organized as follows. In Sec.~\ref{sec:relwork}, we review state-of-the-art approaches for instance segmentation, focusing on methods designed for robotics. Then, in Sec.~\ref{sec:methods}, we describe the proposed training pipeline for fast learning of instance segmentation. In Sec.~\ref{sec:exp_setup}, we report on the experimental setup used to validate our approach. We then benchmark our approach on the two considered robotics datasets in Sec.~\ref{sec:results}. In Sec.~\ref{sec:rpn}, we specifically quantify the benefit of the adaptation of the region proposal. In Sec.~\ref{sec:towards_robot}, we simulate the robotic scenario in which data come into stream and we discuss various performance trade-offs. Then, in Sec.~\ref{sec:robotics_application}, we describe an incremental version of the proposed pipeline and we deploy it on a robotic platform. Finally, in Sec.~\ref{sec:conclusions} we draw conclusions.

\section{RELATED WORK}
\label{sec:relwork}
In this section, we provide an overview of state-of-the-art methods for instance segmentation (Sec.~\ref{rel_work:segm}), focusing on their application in robotics (Sec.~\ref{rel_work:robotics}).

\subsection{Instance Segmentation}
\label{rel_work:segm}

Approaches proposed in the literature to address instance segmentation can be classified in the following three groups.

\noindent
\textbf{Detection-based instance segmentation.} Methods in this category extend approaches for object detection, by adding a branch for mask prediction within the bounding boxes proposed by the detector. Therefore, as for object detection methods, they can be grouped in \textit{(i)} \textit{multi-stage} (also known as \textit{region-based}) and \textit{(ii)} \textit{one-stage}. Methods from the first group rely on detectors that firstly predict a set of candidate regions and then classify and refine each of them (e.g. Faster R-CNN~\cite{ren2015_faster} or R-FCN~\cite{dai2016}).
\textit{One-stage} detectors, instead, solve the object detection task in one forward pass of the network. Differently from \textit{multi-stage} approaches, they do not perform any per-region operation, like e.g. per-region feature extraction and classification (see for instance, EfficientDet~\cite{tan2020efficientdet} and YOLOv3~\cite{Redmon2018}).

\noindent
The representative method among the \textit{multi-stage} approaches is Mask R-CNN~\cite{He2017} that builds on top of the detection method Faster R-CNN~\cite{ren2015_faster}, by adding a branch for mask prediction (segmentation branch) in parallel to the one for bounding box classification and refinement (detection branch). In Mask R-CNN, input images are initially processed by a convolutional backbone to extract a feature map. This is then used by the Region Proposal Network (RPN) to propose a set of \textit{Regions of Interest} (\textit{RoIs}) that are candidate to contain an object, by associating a class-agnostic objectness score to each region. Then, the \textit{RoI Align} layer associates a convolutional feature map to each \textit{RoI} by \textit{warping} and \textit{cropping} the output of the backbone. These features are finally used for \textit{RoIs} classification, refinement and, subsequently, for mask prediction. In the literature, many other state-of-the-art \textit{multi-stage} approaches for instance segmentation build on top of Mask R-CNN, like Mask Scoring R-CNN~\cite{huang2019mask} or PANet~\cite{liu2018path}.

\noindent
YOLACT~\cite{bolya2019yolact} and BlendMask~\cite{chen2020blendmask} are representative of \textit{one-stage} methods. YOLACT~\cite{bolya2019yolact} extends a backbone RetinaNet-like~\cite{Lin2017focal} detector with a segmentation branch. BlendMask~\cite{chen2020blendmask}, instead, extends FCOS~\cite{tian2019fcos} for mask predictions. An alternative paradigm for instance segmentation based on the \textit{one-stage} detector CenterNet~\cite{zhou2019objects} is Deep Snake~\cite{peng2020}. Differently from the methods mentioned above that predict per-pixel confidence within the proposed bounding boxes, it exploits the circular convolution~\cite{peng2020} to predict an offset for each mask vertex point, starting from an initial coarse contour.

\noindent
\textbf{Labelling pixels followed by clustering.} Approaches in this group build on methods for semantic segmentation, which is the task of classifying each pixel of an image according to its category (being thus agnostic to different object instances). Building on these methods, approaches in the literature separate the different instances by clustering the predicted pixels. As an example, SSAP~\cite{gao2019ssap} uses the so-called \textit{affinity pyramid} in parallel with a branch for semantic segmentation to predict the probability that two pixels belong to the same instance in a hierarchical manner. This is done with the aim of grouping pixels of the same instance. InstanceCut~\cite{kirillov2017instancecut}, instead, exploits an instance-agnostic segmentation and an instance-aware edge predictor to compute the instance-aware segmentation of an image. Finally, the method proposed in~\cite{bai2017} learns the watershed transform with a convolutional neural network, the \textit{Deep Watershed Transform}, given an image and a semantic segmentation. This is done to predict an energy map of the image, where the energy basins represent the object instances. This information is then used, with a cut at a single energy level, to produce connected components corresponding to different object instances.

\noindent
\textbf{Dense sliding window.} These approaches simultaneously predict mask instances and their class-agnostic or class-specific scores. For instance, DeepMask~\cite{pinheiro2016learning} predicts in parallel a class-agnostic mask and an objectness score for each patch of an input image with a shallow convolutional neural network. InstanceFCN~\cite{dai2016instance}, alternatively, predicts an instance sensitive score map for each window of the considered input image. This method exploits local coherence for class-agnostic masks prediction, and, as DeepMask, per-window class-agnostic scores. Similarly, TensorMask~\cite{chen2019tensormask} predicts class-agnostic instance masks, but it leverages on the proposed mask representation as a 4D tensor to preserve the spatial information among pixels. Moreover, the classification branch of the proposed approach outputs a class-specific score, thus improving the class-agnostic predictions provided by DeepMask and InstanceFCN.

\subsection{Instance Segmentation in Robotics}
\label{rel_work:robotics}

The instance segmentation task plays a central role in robotics, not only for providing an accurate 2D scene description for a robot, but also to support other tasks like 6D object pose estimation~\cite{xiang2018posecnn} or computation of grasp candidates~\cite{8506350}. In the literature, the problem is tackled in different ways, depending on the target application. In~\cite{wada2019joint} and~\cite{li2020one} the problem is addressed in cluttered scenarios, while~\cite{li2020learning} and~\cite{danielczuk2019segmenting} propose adopting synthetic data (both images and depth information) for training. In this work, instead, we focus on learning to segment previously unseen objects. In the following paragraphs, we will cover the main literature on this topic.

Some works propose to generalize to unseen objects in a class-agnostic fashion. However, these methods either focus on particular environments, such as tabletop settings, as in~\cite{xie2020best} and~\cite{xie2021unseen}, or require some post-processing~\cite{kuo2019shapemask} which may be unfeasible during the robot operation.

Approaches as the ones proposed in~\cite{pathak2018learning} and~\cite{eitel2019self} 
learn to segment new objects instances by interacting with them. Nevertheless, similarly to the class-agnostic approaches, they are constrained to tabletop settings.

The latest literature on \textit{Video Object Segmentation} provides some methods for learning to segment a set of previously unseen objects in videos. They deal with the problem either in a semi-supervised way~\cite{DAVIS2020-Semi-Supervised-1st}, leveraging on the ground-truth masks of the objects in the first frame of the video, or in an unsupervised fashion~\cite{DAVIS2020-Unsupervised-1st}. They allow to learn to segment new object instances in a shorter time than that required by the fully supervised approaches presented in Sec.~\ref{rel_work:segm}. They typically rely on pre-training a network for instance segmentation and on the subsequent fine-tuning on the target video sequence frames~\cite{voigtlaender2017online}. Some of these approaches have been targeted for robotic scenarios. For instance, the method in~\cite{siam2019video} proposes to learn to segment novel objects in a \textit{Human-Robot Interaction (HRI)} application, leveraging only on objects motion cues. Nevertheless, these approaches are known to suffer from changes of the objects appearance through the video sequence and error drifts~\cite{DAVIS2020-Semi-Supervised-1st}. 

We instead focus on learning to segment novel objects in a class-specific fashion, keeping the performance provided by the state-of-the-art but reducing the required training time. All the approaches mentioned in Sec.~\ref{rel_work:segm} rely on convolutional neural networks that require to be trained end-to-end via backpropagation and stochastic gradient descent. Despite providing impressive performance, they require long training time and large amounts of labeled images to be optimized. These constraints make the adoption of such approaches in robotics difficult, especially for robots operating in unconstrained environments, that require fast adaptation to new objects.

Incremental learning aims at learning new objects instances without degrading performance on the previously known classes. Nevertheless, these approaches rarely focus on speeding-up the training of the models, which may be crucial in robotic applications. Moreover, the current literature in this field mainly focuses on object recognition~\cite{7989364, MALTONI201956}, object detection~\cite{shmelkov2017incremental, perez2020incremental} or semantic segmentation problems~\cite{Michieli_2019_ICCV}, while we target an instance segmentation application. As we show in Sec.~\ref{sec:robotics_application}, we deploy the proposed pipeline on the iCub humanoid robot, adapting it to an incremental setting, where the target classes are not known a-priori.

In this work, we propose a pipeline and a training protocol for instance segmentation which is specifically designed to reduce training time, while preserving performance as much as possible. This approach is based on Mask R-CNN~\cite{He2017}, in which the final layers of the RPN and of the detection and segmentation branches have been replaced with ``shallow'' classifiers based on a fast Kernel-based method optimized for large scale problems~\cite{falkon2018, falkonlibrary2020}. The backbone of the network is trained off-line, while the Kernel-based classifiers are adapted on-line. In this paper, we build on our previous work~\cite{ceola2020segm}, in that we include the adaptation of the region proposal network and a novel training protocol which allows to further reduce the training time. This makes the pipeline suitable for on-line implementation.

\section{METHODS}
\label{sec:methods}
\begin{figure*}
	\centering
	\includegraphics[width=0.94\linewidth]{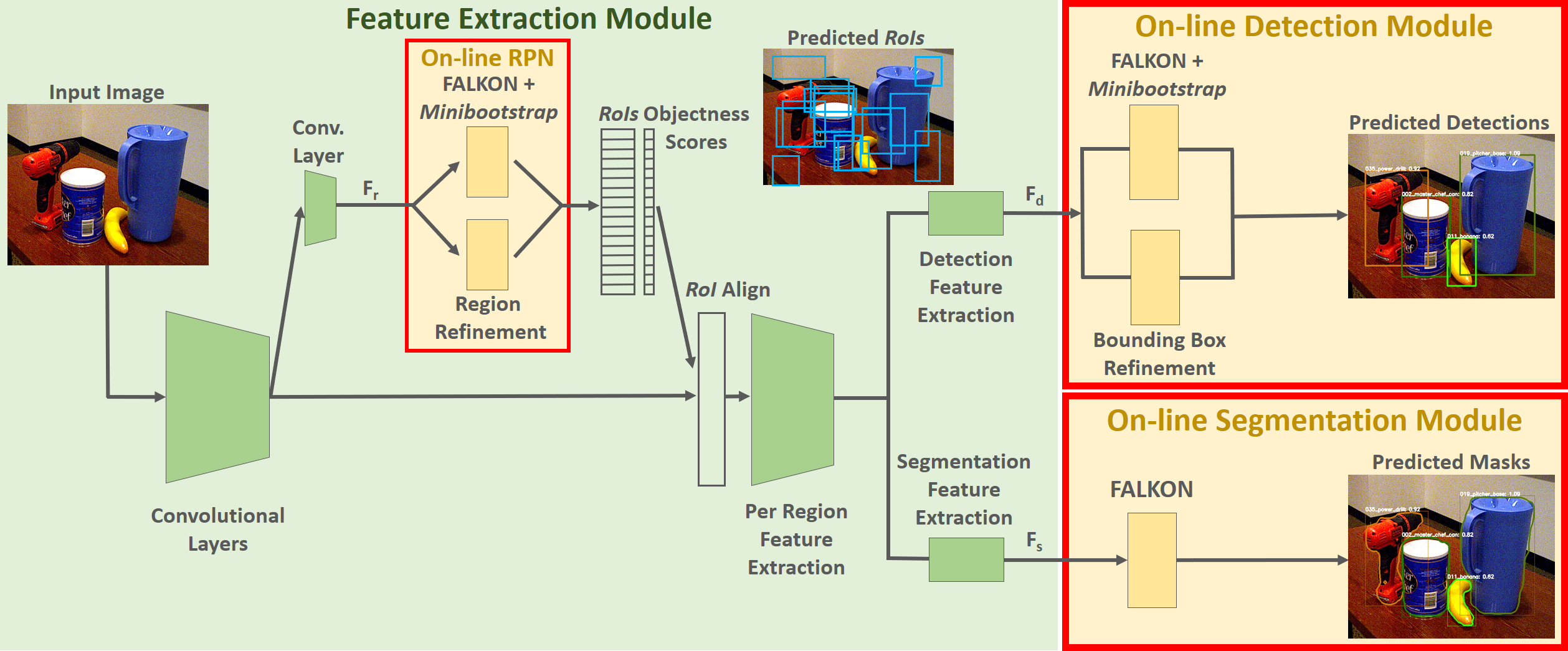}
	\caption{\textbf{Overview of the proposed pipeline}. The \textit{Feature Extraction Module} is composed of Mask R-CNN's first layers trained off-line on the FEATURE-TASK. The three sets of features for \textit{(i)} region proposal ($F_r$), \textit{(ii)} object detection ($F_d$) and \textit{(iii)} instance segmentation ($F_s$) are fed to \textit{(i)} the \textit{On-line RPN}, \textit{(ii)} the \textit{On-line Detection Module} and \textit{(iii)} the \textit{On-line Segmentation Module}.
	At inference time, we substitute the final layers of the Mask R-CNN's RPN with the \textit{On-line RPN} trained on the TARGET-TASK and, as in Mask R-CNN, the output of the \textit{On-line Detection Module} is fed as input to the \textit{RoI Align} to compute the objects masks within the proposed bounding boxes.}
	\label{fig:pipeline}
\end{figure*}

The proposed hybrid pipeline allows to quickly learn to predict masks of previously unseen objects (TARGET-TASK). We rely on convolutional weights pre-trained on a different set of objects (FEATURE-TASK) and we rapidly adapt three modules for region proposal, object detection and mask prediction on the new task. This allows to achieve on-line adaptation on novel objects and visual scenarios.

\subsection{Overview of the Pipeline}
\label{methods:overview}

The proposed pipeline is composed of four modules, which are depicted in Fig.~\ref{fig:pipeline}. They are:
\begin{itemize}
    \item \textbf{Feature Extraction Module.} This is composed of the first layers of Mask R-CNN, which has been pre-trained off-line on the FEATURE-TASK. We use it to extract the convolutional features to train the three on-line modules on the TARGET-TASK. In particular, we use it to extract the features  $F_r$, $F_d$ and $F_s$ from the penultimate layers of the RPN, and of the detection and segmentation branches, respectively.
    \item \textbf{On-line RPN.} This replaces the last layers of the Mask R-CNN's RPN to predict a set of regions that likely contain an object in an image, given a feature map $F_r$. We describe the training procedure in Sec.~\ref{methods:bbox_learning}.
    \item \textbf{On-line Detection Module.} This is composed of classifiers and regressors that, starting from a set of feature tensors $F_d$, classify and refine the regions proposed by the \textit{On-line RPN}. See Sec.~\ref{methods:bbox_learning} for the description of the training procedure.
    \item \textbf{On-line Segmentation Module.} Given a feature map $F_s$, this module predicts the masks of the objects within the detections proposed by the \textit{On-line Detection Module}. We describe the training procedure in Sec.~\ref{methods:oos_learning}.
\end{itemize}
In the three on-line modules described above, we use FALKON for classification. This is a Kernel-based method optimized for large-scale problems~\cite{falkon2018}. In particular, we use the implementation available in~\cite{falkonlibrary2020}.

\subsection{Bounding Box Learning}
\label{methods:bbox_learning}

The prediction of region proposal candidates and object detection are problems that share similarities. In both cases, input bounding boxes are classified and then refined, starting from associated feature tensors as input. In our pipeline, these problems are carried out by the \textit{On-line RPN} and the \textit{On-line Detection Module}, which are implemented by $N_c$ FALKON binary classifiers and $4N_c$ Regularized Least Squares (RLS) regressors~\cite{girshick2014_rcnn}. Specifically, $N_c$ represents: \textit{(i)} the number of anchors for the \textit{On-line RPN} (see the following paragraphs) or \textit{(ii)} the number of semantic classes of the TARGET-TASK for the \textit{On-line Detection Module}. In both the on-line modules, we tackle the well known problem of foreground-background imbalance of training samples in object detection~\cite{Lin2017focal} by adopting the \textit{Minibootstrap} strategy proposed in~\cite{maiettini2018, maiettini2019a} for FALKON training. The \textit{Minibootstrap} is an approximated procedure for hard negatives mining~\cite{Sung1996,girshick2014_rcnn} that allows to iteratively select a subset of hard negative samples to balance the training sets associated to each of the $N_c$ classes. We report the pseudo-code of the \textit{Minibootstrap} procedure in App.~\ref{appendix:minibootstrap}. The RLS regressors for boxes refinement, instead, are trained on a set of positive (foreground) instances.

In the \textit{On-line RPN}, the classifiers are trained on a binary task to discriminate anchors representing the background from those representing \textit{RoIs}, i.e., containing an instance of any of the TARGET-TASK classes. An anchor~\cite{ren2015_faster} is a bounding box of a predefined size and aspect ratio centered on an image pixel. For each pixel, there are a fixed number of anchors of different form factors and one classifier is instantiated for each of them. In the \textit{On-line Detection}, instead, a binary classifier is instantiated for each class. Each classifier is trained to discriminate regions proposed by the \textit{On-line RPN} depicting an object of its class from other classes or background.

\begin{figure}
	\centering
	\includegraphics[width=1.0\linewidth]{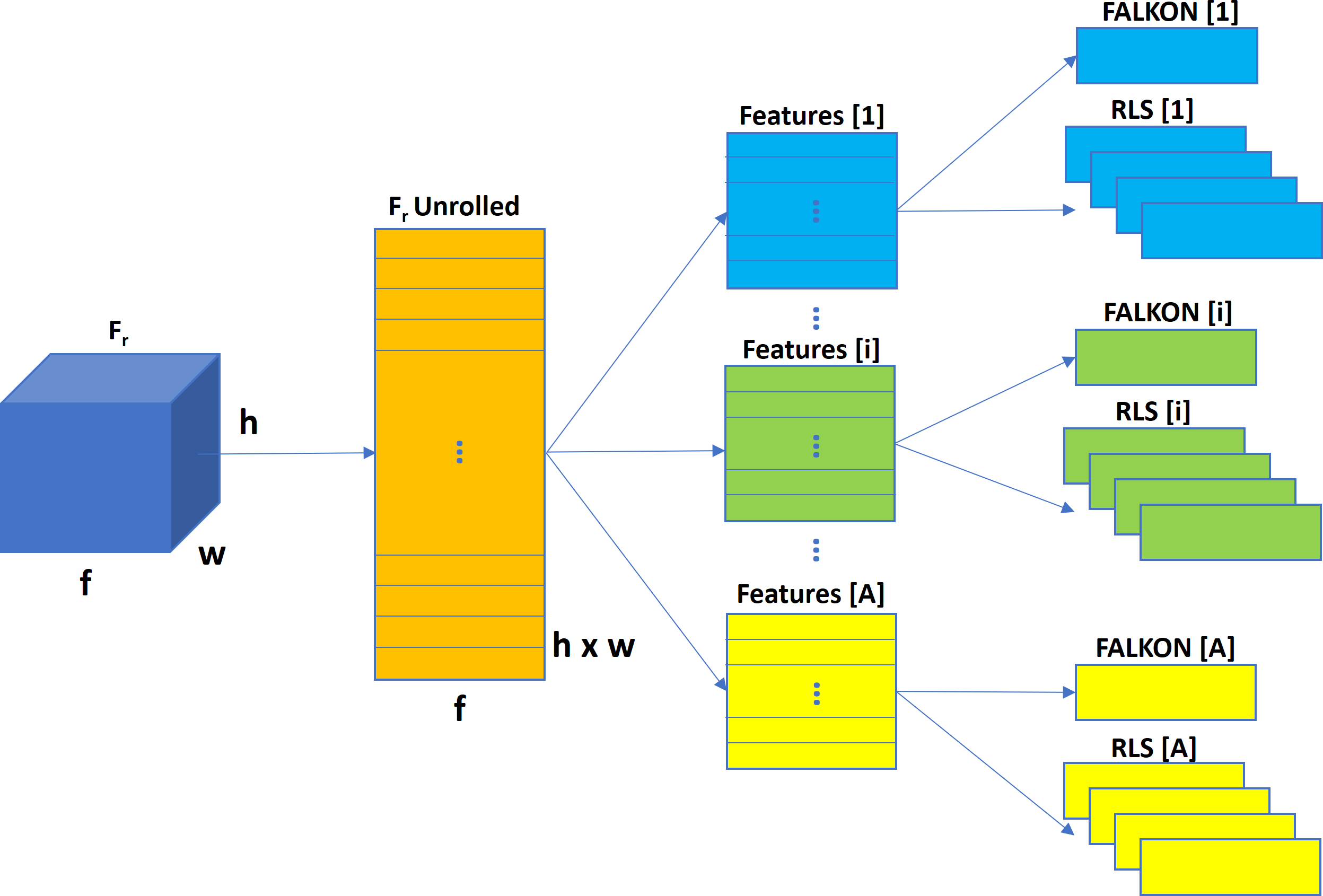}
	\caption{\textbf{On-line RPN.} Given the feature map $F_r$, this is unrolled into $h{\times}w$ tensors of features of size $f$ ($F_r$ \textit{Unrolled}). A subset of these features is chosen to train a FALKON classifier and four RLS regressors for each anchor.}
	\label{fig:rpn}
\end{figure}

\noindent
\textbf{On-line RPN.} In Mask R-CNN's RPN, the classification is performed on a set of anchors $A$ (i.e., $N_c=A$). Given the input feature map computed by the backbone of height $h$, width $w$ and with $f$ channels, the RPN firstly processes it with a convolutional layer to obtain a feature map $F_r$ of the same size ($h{\times}w{\times}f$). Then, $F_r$ is processed by two convolutional layers. One is composed of $A$ convolutional kernels which compute the objectness scores of each considered anchor. This layer computes an output tensor of size $h{\times}w{\times}A$, in which the $ija^{th}$ element is the objectness score of the $a^{th}$ anchor in the location $(i,j)$. The other output layer, instead, is composed of $4A$ kernels for the refinement of such bounding boxes. It computes $h{\times}w{\times}4A$ values for the refinement of the regions associated to the anchors at each location $(i,j)$. Both the output convolutional kernels have size $1$ and stride $1$.\\
As shown in Fig.~\ref{fig:rpn}, we replace the $A$ convolutional kernels for the computation of the objectness scores with $A$ FALKON binary classifiers and we train them with the \textit{Minibootstrap}. We use the $h\times w$ tensors of size $f$ resulting from the flattening of the feature maps $F_r$ as training features. The considered positive features are those associated to a specific location of an anchor whose \textit{Intersection over Union} (\textit{IoU}) with at least a ground-truth bounding box is greater than $0.7$\footnote{For the \textit{On-line RPN}, we set the positive and negative thresholds for the classifiers as in Mask R-CNN's RPN~\cite{He2017}.} (in case there are no anchors overlapping the ground-truth bounding boxes with \textit{IoU} $>0.7$, the ones with the highest \textit{IoU} are chosen as positives). The feature tensors for the background, instead, are those whose \textit{IoU} with the ground-truths is smaller than $0.3$. Similarly, we replace the $4A$ convolutional kernels for the refinement of the proposed regions with $4A$ RLS regressors ($4$ RLS for each anchor). We consider as training samples for the $4$ regressors associated to each anchor a set of features chosen as the positive samples for the FALKON classifiers, but setting the \textit{IoU} threshold to $0.6$\footnote{\label{rls_fn}We set the value of the \textit{IoU} threshold for the RLS regressors as in R-CNN~\cite{girshick2014_rcnn}.}.

\begin{figure}
	\centering
	\includegraphics[width=1.0\linewidth]{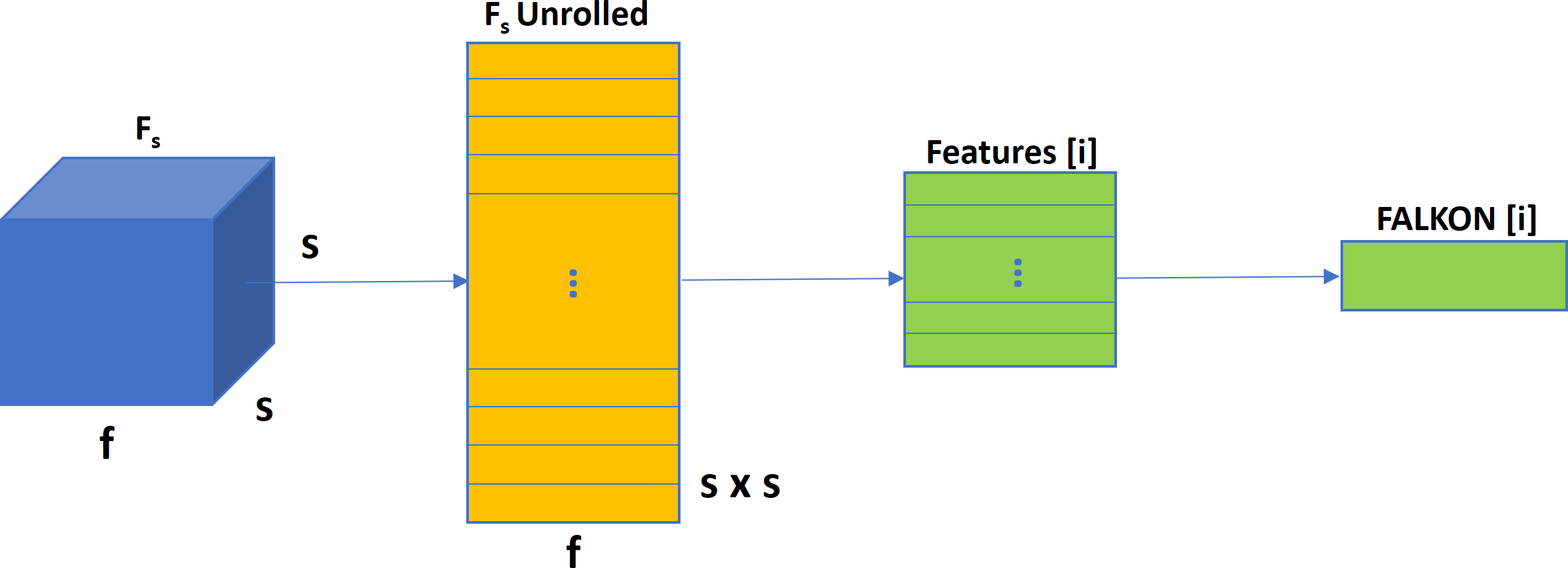}
	\caption{\textbf{On-line Segmentation.} Given the feature map $F_s$ associated to a \textit{RoI} of class \textit{i}, this is unrolled into $s{\times}s$ tensors of features of size $f$ ($F_s$ \textit{Unrolled}) from which positive and negative features are sampled to train the $i^{th}$ FALKON per-pixel classifier. Note that this procedure is performed for each \textit{RoI} of the $N$ classes.}
	\label{fig:oos}
\end{figure}
\noindent
\textbf{On-line object detection.} We train the \textit{On-line Detection Module} with the strategy illustrated above, considering the $N$ classes of the TARGET-TASK (i.e., $N_c=N$). As training samples, we consider the tensors of features produced by the penultimate layer of the Mask R-CNN's detection branch ($F_d$) associated to each \textit{RoI} proposed by the region proposal method. In particular, we consider as positive samples for the $n^{th}$ FALKON classifier, those \textit{RoIs} with \textit{IoU} $>0.6$\footnote{We consider as positive samples for the classifiers in the \textit{On-line Detection Module} the training features for region refinement as in~\cite{He2017}.} with a ground-truth box of an instance of class $n$ ($n \in [1,\dots, N]$). The same positive samples are also used for training the $n^{th}$ RLS regressor\footref{rls_fn}. Then, as negative samples, we consider the \textit{RoIs} with \textit{IoU} $<0.3$\footnote{For the classifiers in the \textit{On-line Detection Module} we define the negative samples as in~\cite{girshick2014_rcnn}.} with the ground-truths of class $n$.

\begin{figure*}
	\centering
	\includegraphics[width=1.0\linewidth]{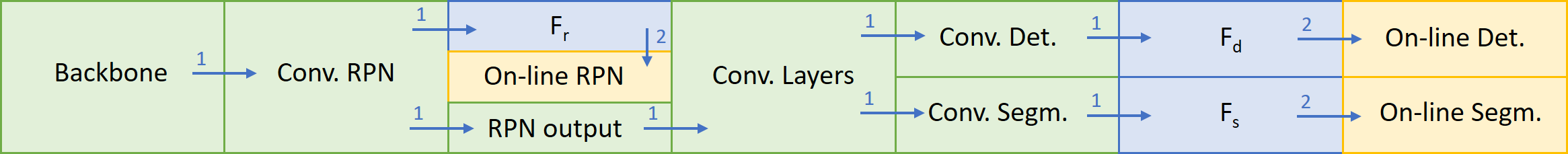}
	\caption{\textbf{Ours} training protocol. We rely on the feature extraction layers of Mask R-CNN pre-trained on the FEATURE-TASK to simultaneously extract $F_r$, $F_d$ and $F_s$. We then use these features to train the three on-line modules on the TARGET-TASK. The values on the arrows correspond to the training steps in Sec.~\ref{methods:training_protocols}.}
	\label{fig:indep}
\end{figure*}

\subsection{On-line Segmentation}
\label{methods:oos_learning}

In Mask R-CNN, in the configuration that does not use the \textit{Feature Pyramid Network} (FPN)~\cite{Lin2017_fpn} in the backbone, the segmentation branch is a shallow fully convolutional network (FCN) composed of two layers that takes as input a feature map of size $s{\times}s{\times}f$ associated to each \textit{RoI}. The first layer processes the input feature map into another feature map $F_s$ of the same size. The last convolutional layer, instead, has $N$ channels (one for each class of the TARGET-TASK) and kernel size $1$ and stride $1$. Therefore, the output of the Mask R-CNN's segmentation branch is a tensor of size $s{\times}s{\times}N$, where the $ijn^{th}$ value of such tensor represents the confidence that the pixel in the location $(i,j)$ of the \textit{RoI} corresponds to the $n^{th}$ class.

For the fast learning of the \textit{On-line Segmentation Module}, we rely on the first layer of the segmentation branch for feature extraction, but we substitute the last convolutional layer for per-pixel prediction with $N$ FALKON binary classifiers. To train such classifiers, we consider the ground-truth boxes of each training image, we compute the feature map of size $s{\times}s{\times}f$ for each of them and we flatten each of such feature maps into $s{\times}s$ tensors of size $f$, as shown in Fig.~\ref{fig:oos}. Among these tensors, we consider as positive samples for the $n^{th}$ classifier the features associated to the pixels in the ground-truth masks of class $n$. Instead, we consider as negative samples the features associated to the background pixels contained in ground-truth bounding boxes of class $n$. Given the great amount of training samples, to speed-up the training procedure, we randomly subsample both the positive and the negative features by a factor $r$. According to the analysis provided in~\cite{ceola2020segm}, we set $r$ to $0.3$.

\subsection{Training Protocol}
\label{methods:training_protocols}
 
In this work, we propose a training protocol that allows to quickly update
the \textit{On-line RPN}, the \textit{On-line Detection Module} and the \textit{On-line Segmentation Module}.
The proposed method (referred to as \textbf{Ours}) starts with the weights of Mask R-CNN pre-trained on the FEATURE-TASK and adapts the on-line modules on the TARGET-TASK. This is composed of the two steps depicted in Fig.~\ref{fig:indep}:

\begin{enumerate}
    \item \textit{Feature extraction.} This is done with a forward pass of the pre-trained Mask R-CNN feature extractor to compute $F_r$, $F_d$ and $F_s$.
    \item \textit{On-line training.} The set of features $F_r$, $F_d$ and $F_s$ are used to respectively train \textit{(i)} the \textit{On-line RPN}, \textit{(ii)} the \textit{On-line Detection Module} and \textit{(iii)} the \textit{On-line Segmentation Module} on the TARGET-TASK.
\end{enumerate}

The training features fed to the \textit{On-line Detection Module} are those associated to the regions proposed by the Mask R-CNN's RPN pre-trained on the FEATURE-TASK. These regions are different (and therefore sub-optimal) with respect to the ones that would be proposed by a region proposal method that has been adapted on the TARGET-TASK. Training the \textit{On-line Detection Module} using features extracted from the \textit{On-line RPN} after its adaptation is possible. This, however, would require two feature extraction steps (one for training the \textit{On-line RPN} and the other to train the \textit{On-line Detection Module} and the \textit{On-line Segmentation Module}), which is computationally expensive. For this reason, we consider the \textit{On-line Detection Module} obtained with \textbf{Ours} as an approximation of the one that would be provided by the serial training (see Sec.~\ref{sec:rpn:indep}). Instead, the adaptation of the \textit{On-line Segmentation Module} is not affected by this approximation, since we sample the training features for this module from the ground-truth bounding boxes.

While this approximation is a key component of the proposed training protocol, at inference time features fed to the \textit{On-line Detection Module} and to the \textit{On-line Segmentation Module} are those associated to the regions proposed by the \textit{On-line RPN} trained on the TARGET-TASK, as depicted in Fig.~\ref{fig:pipeline}.

In this work, we show that a single feature extraction step can be performed paying a small price in terms of accuracy (see Sec.~\ref{sec:rpn:indep}), allowing to further improve the training in the on-line implementation (see Sec.~\ref{sec:towards_robot} and Sec.~\ref{sec:robotics_application}).

\section{EXPERIMENTAL SETUP}
\label{sec:exp_setup}
In this section, we report on the experimental settings that we employ for validating the proposed approach. We first evaluate our approach in an off-line setting (Sec.~\ref{sec:results}, \ref{sec:rpn} and \ref{sec:towards_robot}), analyzing performance on two different robotics datasets. Then, we validate it in a real robotic application (Sec.~\ref{sec:robotics_application}), i.e., in an on-line setting. Therefore, in this section, we firstly report on the off-line experimental setup (Sec.~\ref{sec:exp_setup:offline}) and the datasets (Sec.~\ref{sec:exp_setup:datasets}) that we use in our experiments. Then, in Sec.~\ref{sec:exp_setup:robot}, we describe the settings considered for the deployment in a real robotic scenario.

\subsection{Off-line Experiments}
\label{sec:exp_setup:offline}

For our experiments, we compare the proposed method, \textbf{Ours}, with two Mask R-CNN~\cite{He2017} baselines. In particular, we consider:
\begin{itemize}
    \item \textbf{Mask R-CNN (output layers)}: starting from the Mask R-CNN weights pre-trained on the FEATURE-TASK, we re-initialize the output layers of the RPN and of the detection and segmentation branches, and we fine-tune them on the TARGET-TASK, freezing all the other weights of the Mask R-CNN network.
    \item \textbf{Mask R-CNN (full)}: we use the weights of Mask R-CNN pre-trained on the FEATURE-TASK as a warm-restart to train Mask R-CNN on the TARGET-TASK.
\end{itemize}
Specifically, we rely on Mask R-CNN~\cite{He2017}, using ResNet-50~\cite{He2015} as backbone, for the feature extraction of \textbf{Ours} and for the baselines. In App.~\ref{appendix:summary_tab}, we report a summary of the training protocols used in this work.

In all cases, we choose hyper-parameters providing the highest performance on a validation set. Specifically, for \textbf{Ours} we cross-validate the standard deviation of FALKON's Gaussian kernels (namely, $\sigma$) and FALKON's regularization parameter (namely, $\lambda$) for the \textit{On-line RPN}, the \textit{On-line Detection Module} and the \textit{On-line Segmentation Module}. Regarding the baselines, instead, for the experiments in Sec.~\ref{sec:results} and~\ref{sec:rpn}, we train \textbf{Mask R-CNN (output layers)} and \textbf{Mask R-CNN (full)} for the number of epochs that provides the highest segmentation accuracy on the validation set.

For \textbf{Ours}, we set the number of Nystr{\"o}m centers $M$ of the FALKON classifiers composing the \textit{On-line RPN}, the \textit{On-line Detection Module} and the \textit{On-line Segmentation Module} to $1000$, $1000$ and $500$, respectively. Moreover, to train both the \textit{On-line RPN} and the \textit{On-line Detection Module}, we set to $2000$ the size of the batches, $BS$, considered in the \textit{Minibootstrap}.

\noindent
\textbf{Evaluation metrics.} We consider the \textit{mean Average Precision (mAP)} as defined in~\cite{pascal2010} for both object detection and segmentation. Specifically, the accuracy of the predicted bounding boxes will be referred to as \textbf{mAP bbox(\%)} and the accuracy of the mask instances as \textbf{mAP segm(\%)}. For each of them, we consider as positive matches the bounding boxes and the masks whose \textit{IoU} with the ground-truths is greater or equal to a threshold. In our experiments we consider two different thresholds to evaluate different levels of accuracy, namely, $50\%$ (\textbf{mAP50}) and $70\%$ (\textbf{mAP70}). In App.~\ref{appendix:summary_tab} we overview the acronyms considered for \textit{mAP} computation. We also evaluate the methods on the time required for training\footnote{All the \textit{off-line experiments} have been performed on a machine equipped with Intel(R) Xeon(R) W-2295 CPU @ 3.00GHz, and a single NVIDIA Quadro RTX 6000.}. For the Mask R-CNN baselines, the training time is the time needed for their optimization via backpropagation and stochastic gradient descent. As regards \textbf{Ours}, instead, except where differently specified, it is the time necessary for extracting the features and training the on-line modules. For each experiment, we run three trials for each method. We report results in terms of average and standard deviation of the accuracy and of average training time.

\subsection{Datasets}
\label{sec:exp_setup:datasets}
In our experiments we consider three datasets. Specifically, we use MS COCO~\cite{coco} as FEATURE-TASK and the two datasets YCB-Video~\cite{xiang2018posecnn} and HO-3D~\cite{hampali2020honnotate} as TARGET-TASKs to validate our approach. We opted to validate our system on these datasets, which are composed of streams of frames in tabletop and hand-held settings, to be close to our target application. These datasets are usually considered for the task of 6D object pose estimation, however, they are annotated also with object masks. Specifically:
\begin{itemize}
    \item \textbf{MS COCO}~\cite{coco} is a general-purpose dataset, which contains 80 objects categories, for object detection and segmentation. 
    \item \textbf{YCB-Video}~\cite{xiang2018posecnn} is a dataset for 6D pose estimation in which 21 objects from the YCB~\cite{calli2015ycb} dataset are arranged in cluttered tabletop scenarios, therefore presenting strong occlusions. It is composed of video sequences where the tabletop scenes are recorded under different viewpoints. We use as training images a set of 11320 images, obtained by extracting one image every ten from the total 80 training video sequences. As test set, instead, we consider the 2949 \textit{keyframe}~\cite{xiang2018posecnn} images chosen from the remaining 12 sequences. For hyper-parameters cross-validation, we randomly select a subset of 1000 images from the 12 test sequences, excluding the \textit{keyframe} set.
    \item \textbf{HO-3D}~\cite{hampali2020honnotate} is a dataset for hand-object pose estimation, in which objects are a subset of the ones in \textbf{YCB-Video}. It is composed of video sequences, in which a moving hand-held object is shown to a fixed camera. For choosing the training and test sets, we split the available annotated sequences in HO-3D\footnote{Note that, in HO-3D, the annotations for instance segmentation are not provided for the test set. Therefore, we extract training and test sequences from the original HO-3D training set.}, such that, we gather one and at most four sequences for testing and training, respectively. In particular, we use 20156 images as training set, which result from the selection of one every two images from 34 sequences. Instead, we consider as test set 2020 images chosen one every five frames taken from other 9 sequences. For hyper-parameters cross-validation, we consider 2160 frames chosen one every five images, from a subset of 9 sequences taken from the training set (see App.~\ref{appendix:ho3d} for further details).
\end{itemize}

\subsection{Robotic Setup}
\label{sec:exp_setup:robot}

We deploy the proposed pipeline for on-line instance segmentation on the humanoid robot iCub\footnote{We run the module with the proposed method on a machine equipped with Intel(R) Core(TM) i7-9750H CPU @ 2.60GHz, and a single NVIDIA RTX 2080 Ti.}~\cite{icub}. It is equipped with a \textit{Intel(R) RealSense D415} on a headset for the acquisition of RGB images and depth information. We rely on the YARP~\cite{Mettayarp} middleware for the implementation and the communication between the different modules (see Sec.~\ref{sec:robotics_application}). With the exception of the proposed one, we rely on publicly available modules\footnote{\url{https://github.com/robotology}}. We set all the training hyper-parameters as described in Sec.~\ref{sec:exp_setup:offline}.

\section{RESULTS}
\label{sec:results}
\begin{table*}[]
	\centering
	\begin{tabular}{c|c|c|c|c|c}
		\cline{1-6}
		\centering \textbf{Method}                                                  & \textbf{mAP50 bbox(\%)}    & \textbf{mAP50 segm(\%)}   & \textbf{mAP70 bbox(\%)}   & \textbf{mAP70 segm(\%)}   & \textbf{Train Time}   \\ \hline
		\multicolumn{1}{c|}{Mask R-CNN (full)}          		                            & 89.66 $\pm$ 0.47                      & 91.26 $\pm$ 0.56                       & 84.67 $\pm$ 0.81                       & 80.26 $\pm$ 0.59                        & 1h 35m 42s                   \\ \hline 
		\multicolumn{1}{c|}{Mask R-CNN (output layers)}          		                    & 84.51 $\pm$ 0.40                      & 81.70 $\pm$ 0.17                       & 75.81 $\pm$ 0.30                       & 70.46 $\pm$ 0.24                        & 2h 57m 12s                   \\ \hline \rowcolor[HTML]{79A6D2} 
		\multicolumn{1}{c|}{\textbf{Ours}}          		                                    & 83.66 $\pm$ 0.84                      & 83.06 $\pm$ 0.92                       & 72.97 $\pm$ 1.02                       & 68.11 $\pm$ 0.29                          & 13m 53s                      \\ \hline
	\end{tabular}
	\caption[]{Benchmark on the YCB-Video dataset. We compare the proposed approach \textbf{Ours} to the baseline \textbf{Mask R-CNN (output layers)} and to the upper bound \textbf{Mask R-CNN (full)}.}
	\label{table:bench_ycbv}
\end{table*}

\begin{table*}[]
	\centering
	\begin{tabular}{c|c|c|c|c|c}
		\cline{1-6}
		\centering \textbf{Method}                                                  & \textbf{mAP50 bbox(\%)}    & \textbf{mAP50 segm(\%)}   & \textbf{mAP70 bbox(\%)}   & \textbf{mAP70 segm(\%)}   & \textbf{Train Time}   \\ \hline
		\multicolumn{1}{c|}{Mask R-CNN (full)}          		                            & 92.21 $\pm$ 0.88                      & 90.70 $\pm$ 0.17                       & 86.73 $\pm$ 0.71                       & 77.25 $\pm$ 0.62                        & 38m 38s                   \\ \hline
		\multicolumn{1}{c|}{Mask R-CNN (output layers)}          		                    & 88.05 $\pm$ 0.32                      & 86.11 $\pm$ 0.29                       & 74.75 $\pm$ 0.19                       & 65.04 $\pm$ 0.62                        & 1h 50m 33s                   \\ \hline
		\rowcolor[HTML]{79A6D2} 
		\multicolumn{1}{c|}{\textbf{Ours}}          		                                    & 83.63 $\pm$ 1.64                      & 84.50 $\pm$ 1.63                       & 63.33 $\pm$ 1.65                       & 61.54 $\pm$ 0.33                        & 16m 51s                   \\ \hline
	\end{tabular}
	\caption{Benchmark on the HO-3D dataset. We report on the performance obtained with \textbf{Ours} and we compare it to \textbf{Mask R-CNN (output layers)} and \textbf{Mask R-CNN (full)} for the analysis in Sec.~\ref{sec:results:ho3d}.}
	\label{table:bench_ho3d}
\end{table*}

In this section, we benchmark the proposed approach on YCB-Video (Sec.~\ref{sec:results:ycbv}) and HO-3D (Sec.~\ref{sec:results:ho3d}).

\subsection{Benchmark on YCB-Video}
\label{sec:results:ycbv}

We consider the 21 objects from YCB-Video as TARGET-TASK and we compare the performance of \textbf{Ours} against the baseline \textbf{Mask R-CNN (output layers)}. We also report the performance of \textbf{Mask R-CNN (full)}, which can be considered as an upper-bound because, differently from the proposed method, it updates both the feature extraction layers and the output layers (i.e., the backbone, the RPN and the detection and segmentation branches) fitting more the visual domain of the TARGET-TASK. In \textbf{Ours}, we empirically set the number of batches in the \textit{Minibootstrap} to $10$, to achieve the best training time/accuracy trade-off (see Fig.~\ref{fig:online_vs_mask_ycbv} for details).

Results in Tab.~\ref{table:bench_ycbv} show that \textbf{Ours} achieves similar performance as \textbf{Mask R-CNN (output layers)} in a fraction ($\mathbf{{\sim}12.8\times}$ smaller) of the training time. 
In comparison with the upper bound, \textbf{Ours} is not as accurate as \textbf{Mask R-CNN (full)} ($\mathbf{{\sim}9.0\%}$ less precise if we consider the \textbf{mAP50 segm(\%)}), but is trained $\mathbf{{\sim}6.9\times}$ faster.

\subsection{Benchmark on HO-3D}
\label{sec:results:ho3d}

We evaluate the proposed approach on the HO-3D dataset. As in Sec.~\ref{sec:results:ycbv}, we compare \textbf{Ours} with \textbf{Mask R-CNN (output layers)} and we consider \textbf{Mask R-CNN (full)} as upper bound. For this experiment, we empirically set the number of the \textit{Minibootstrap} batches of the \textit{On-line RPN} and of the \textit{On-line Detection Module} in \textbf{Ours} to $12$ and we report the obtained results in Tab.~\ref{table:bench_ho3d}. 

Similarly to the experiment on YCB-Video, \textbf{Ours} can be trained $\mathbf{{\sim}2.3\times}$ and $\mathbf{{\sim}6.6\times}$ faster than \textbf{Mask R-CNN (full)} and \textbf{Mask R-CNN (output layers)}, respectively. Models obtained with \textbf{Ours} are slightly less precise than those provided by \textbf{Mask R-CNN (output layers)} for the task of instance segmentation, while they are $\mathbf{{\sim}15.3\%}$ less accurate if we consider the \textbf{mAP70 bbox(\%)}. We will show in Sec.~\ref{sec:rpn:indep} that this gap can be recovered with a different training protocol (\textbf{Ours Serial} in Sec.~\ref{sec:rpn:indep}). However, \textbf{Ours}, achieves the best training time with an accuracy that is close to the state-of-the-art.

\begin{table*}[]
	\centering
	\begin{tabular}{c|c|c|c|c|c}
		\cline{1-6}
		\centering \textbf{Method}                                                  & \textbf{mAP50 bbox(\%)}    & \textbf{mAP50 segm(\%)}   & \textbf{mAP70 bbox(\%)}   & \textbf{mAP70 segm(\%)}   & \textbf{Train Time}   \\ \hline
		\multicolumn{1}{c|}{Mask R-CNN (full)}          		                            & 89.66 $\pm$ 0.47                      & 91.26 $\pm$ 0.56                       & 84.67 $\pm$ 0.81                       & 80.26 $\pm$ 0.59                        & 1h 35m 42s                   \\ \hline
		\multicolumn{1}{c|}{\textbf{O-OS}}                          		                    & 76.15 $\pm$ 0.31                      & 74.44 $\pm$ 0.11                       & 68.06 $\pm$ 0.34                       & 63.90 $\pm$ 0.36                        & 11m 14s                   \\ \hline
		\rowcolor[HTML]{79A6D2} 
		\multicolumn{1}{c|}{\textbf{Ours}}          		                                    & 83.66 $\pm$ 0.84                      & 83.06 $\pm$ 0.92                       & 72.97 $\pm$ 1.02                       & 68.11 $\pm$ 0.29                          & 13m 53s                      \\ \hline
	\end{tabular}
	\caption[]{Comparison between \textbf{Ours}, \textbf{Mask R-CNN (full)} and \textbf{O-OS} trained on YCB-Video. For \textbf{O-OS}, we reproduce the experiment of Tab.~I in~\cite{ceola2020segm}, but we run the experiment three times (reporting mean and standard deviation of the obtained results) on the hardware used for this work and we set the training hyper-parameters as described in Sec.~\ref{sec:rpn:gain}.}
	\label{table:rpn_ycbv_1}
\end{table*}

\begin{table*}[]
	\centering
	\begin{tabular}{c|c|c|c|c|c}
		\cline{1-6}
		\centering \textbf{Method}                                                  & \textbf{mAP50 bbox(\%)}    & \textbf{mAP50 segm(\%)}   & \textbf{mAP70 bbox(\%)}   & \textbf{mAP70 segm(\%)}   & \textbf{Train Time}   \\ \hline
		\multicolumn{1}{c|}{Mask R-CNN (full)}          		                            & 92.21 $\pm$ 0.88                      & 90.70 $\pm$ 0.17                       & 86.73 $\pm$ 0.71                       & 77.25 $\pm$ 0.62                        & 38m 38s                   \\ \hline
		\multicolumn{1}{c|}{\textbf{O-OS}}                          		                    & 75.27 $\pm$ 0.26                      & 77.42 $\pm$ 0.45                       & 57.89 $\pm$ 0.24                       & 57.86 $\pm$ 0.21                        & 13m 31s                   \\ \hline
		\rowcolor[HTML]{79A6D2} 
		\multicolumn{1}{c|}{\textbf{Ours}}          		                                    & 83.63 $\pm$ 1.64                      & 84.50 $\pm$ 1.63                       & 63.33 $\pm$ 1.65                       & 61.54 $\pm$ 0.33                        & 16m 51s                   \\ \hline
	\end{tabular}
	\caption{We report on the performance obtained on HO-3D with \textbf{Ours} and we compare it to \textbf{Mask R-CNN (full)} and \textbf{O-OS} for the analysis in Sec.~\ref{sec:rpn:gain}.}
	\label{table:rpn_ho3d_1}
\end{table*}

\section{FAST REGION PROPOSAL ADAPTATION}
\label{sec:rpn}
In this section, we investigate the impact of region proposal adaptation on the overall performance. In particular, in Sec.~\ref{sec:rpn:gain}, we show that, with respect to our previous work~\cite{ceola2020segm}, updating the RPN provides a significant gain in accuracy, maintaining a comparable training time. Then, in Sec.~\ref{sec:rpn:indep} we analyze the speed/accuracy trade-off achieved with the proposed approximated training protocol.

\subsection{Is Region Proposal Adaptation Key to Performance?}
\label{sec:rpn:gain}

The adaptation of the region proposal on a new task provides a significant gain in accuracy for object detection (in this paper we report some evidence while additional experiments can be found in~\cite{ceola2020rpn}). In particular, adapting the region proposal is especially effective when FEATURE-TASK and TARGET-TASK present a significant domain shift (which represents a common scenario in robotics). In this section, we show that better region proposals improves also the downstream mask estimation.

For testing performance under domain shift, we consider as FEATURE-TASK the categorization task of the general-purpose dataset MS COCO. Instead, we consider as TARGET-TASKs the identification tasks of the YCB-Video and HO-3D datasets, which depict tabletop and in-hand scenarios, respectively.

We consider \textbf{Mask R-CNN (full)} as the upper bound of the experiment, since it updates the entire network on the new task. We compare \textbf{Ours} with the method proposed in~\cite{ceola2020segm} (Sec.~III), namely \textbf{O-OS}\footnote{In~\cite{ceola2020segm}, \textbf{O-OS} is referred to as \textbf{Ours}.}, in which the RPN remains constant during training on the TARGET-TASK. For a fair comparison, we set \textbf{O-OS} training hyper-parameters according to Sec.~\ref{sec:exp_setup:offline} (i.e., changing the number of Nystr{\"o}m centers of FALKON in the \textit{On-line Detection Module} and in the \textit{On-line Segmentation Module} with respect to~\cite{ceola2020segm}).

Results in Tab.~\ref{table:rpn_ycbv_1} and in Tab.~\ref{table:rpn_ho3d_1} show that, as expected, there is an accuracy gap between \textbf{Mask R-CNN (full)} and all the other considered methods (\textbf{Ours} and \textbf{O-OS}). However, notably, the adaptation of the region proposal on the TARGET-TASK in \textbf{Ours} allows to significantly reduce the accuracy gap between \textbf{Mask R-CNN (full)} and \textbf{O-OS}. Moreover, \textbf{Ours} outperforms the accuracy of \textbf{O-OS} with a comparable training time. For instance, in the HO-3D experiment (see Tab.~\ref{table:rpn_ho3d_1}), the segmentation \textbf{mAP50} obtained with \textbf{Ours} is, on average, $\mathbf{{\sim}7.1}$ points greater than \textbf{O-OS}, with a difference in training time of only \textit{3m 20s}.

\subsection{Approximated On-line Training: Speed/Accuracy Trade-off}
\label{sec:rpn:indep}

\begin{figure*}
	\centering
	\includegraphics[width=1.0\linewidth]{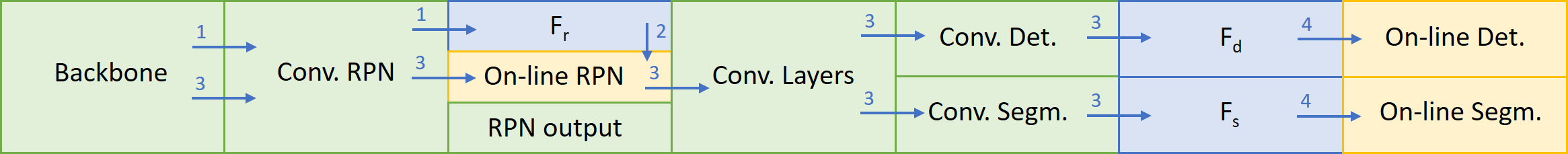}
	\caption{\textbf{Ours Serial} training protocol. We rely on the feature extraction layers of Mask R-CNN pre-trained on the FEATURE-TASK to extract $F_r$ and we train the \textit{On-line RPN} on the TARGET-TASK. Then, we rely on the feature extraction layers of Mask R-CNN and on the \textit{On-line RPN} trained on the TARGET-TASK to extract $F_d$ and $F_s$. Finally, we train the \textit{On-line Detection Module} and the \textit{On-line Segmentation Module} on the TARGET-TASK. The values on the arrows correspond to the training steps in Sec.~\ref{sec:rpn:indep}.}
	\label{fig:serial}
\end{figure*}

In this section, we evaluate the impact of the approximation in \textbf{Ours}. To do this, we compare it with a different training protocol (referred to as \textbf{Ours Serial}). This relies on the same on-line modules as \textbf{Ours}. However, \textbf{Ours Serial} performs two steps of feature extraction, one to train the \textit{On-line RPN} and the other for the \textit{On-line Detection Module} and the \textit{On-line Segmentation Module}. This latter is done after region proposal adaptation and allows to use better \textit{RoIs} to train the module for on-line detection, improving the overall performance of the pipeline. In details, \textbf{Ours Serial} is composed of the four steps depicted in Fig.~\ref{fig:serial}:
\begin{enumerate}
    \item Feature extraction for region proposal. This is done to extract $F_r$ (see Sec.~\ref{methods:overview}) on the images of the TARGET-TASK.
    \item These features are then used to train the \textit{On-line RPN} on the TARGET-TASK, as described in Sec.~\ref{methods:bbox_learning}.
    \item The new \textit{On-line RPN} is used to extract more precise regions and the corresponding features for detection and segmentation (respectively, $F_d$ and $F_s$).
    \item $F_d$ and $F_s$ are used to train the \textit{On-line Detection Module} and the \textit{On-line Segmentation Module} on the TARGET-TASK, as described in Sec.~\ref{methods:bbox_learning} and Sec.~\ref{methods:oos_learning}, respectively.
\end{enumerate}

We evaluate \textbf{Ours} and \textbf{Ours Serial} in the same setting used for previous experiments (Sec.~\ref{sec:results} and Sec.~\ref{sec:rpn:gain}). In \textbf{Ours Serial}, we set the Nystr{\"o}m centers of the FALKON classifiers and the batch size \textit{BS} considered in the \textit{Minibootstrap} to train the \textit{On-line RPN} and the \textit{On-line Detection Module} as described in Sec.~\ref{sec:exp_setup:offline}. Moreover, we empirically set the number of \textit{Minibootstrap} iterations $n_B$ to $8$ and $7$ in the experiments on YCB-Video and HO-3D, respectively. 

We report results in Tab.~\ref{table:rpn_ycbv_2} (YCB-Video) and in Tab.~\ref{table:rpn_ho3d_2} (HO-3D). Specifically, the first one shows that the accuracy of \textbf{Ours} in the YCB-Video experiment is comparable to the one of \textbf{Ours Serial}, demonstrating that the approximated training procedure substantially does not affect performance in this case. Instead, in the HO-3D experiment (see Tab.~\ref{table:rpn_ho3d_2}), \textbf{Ours} is slightly less precise than \textbf{Ours Serial} for the task of instance segmentation, while being $\mathbf{{\sim}11.6\%}$ less accurate if we consider the \textbf{mAP70 bbox(\%)}. However, \textbf{Ours} is trained $\mathbf{{\sim}1.8\times}$ and $\mathbf{{\sim}2.2\times}$ faster than \textbf{Ours Serial} in the YCB-Video and in the HO-3D experiments, respectively.

Still, with respect to \textbf{Mask R-CNN (output layers)}, \textbf{Ours Serial} achieves comparable performance, but with training time that is much shorter. However, the approximated training protocol proposed in this paper allows further optimization which is discussed in the next section.

\begin{table*}[]
	\centering
	\begin{tabular}{c|c|c|c|c|c}
		\cline{1-6}
		\centering \textbf{Method}                                                  & \textbf{mAP50 bbox(\%)}    & \textbf{mAP50 segm(\%)}   & \textbf{mAP70 bbox(\%)}   & \textbf{mAP70 segm(\%)}   & \textbf{Train Time}   \\ \hline
		\multicolumn{1}{c|}{Mask R-CNN (output layers)}          		                    & 84.51 $\pm$ 0.40                      & 81.70 $\pm$ 0.17                       & 75.81 $\pm$ 0.30                       & 70.46 $\pm$ 0.24                        & 2h 57m 12s                   \\ \hline 
		\multicolumn{1}{c|}{\textbf{Ours Serial}}          		                                    & 83.97 $\pm$ 0.59                      & 83.00 $\pm$ 0.78                       & 75.06 $\pm$ 0.88                       & 69.12 $\pm$ 0.56                        & 24m 42s                   \\ \hline
		\rowcolor[HTML]{79A6D2} 
		\multicolumn{1}{c|}{\textbf{Ours}}          		                                    & 83.66 $\pm$ 0.84                      & 83.06 $\pm$ 0.92                       & 72.97 $\pm$ 1.02                       & 68.11 $\pm$ 0.29                          & 13m 53s                      \\ \hline
	\end{tabular}
	\caption[]{Comparison between the proposed approach \textbf{Ours}, the baseline \textbf{Mask R-CNN (output layers)} and \textbf{Ours Serial} trained on YCB-Video. Refer to Sec.~\ref{sec:rpn:indep} for further details.}
	\label{table:rpn_ycbv_2}
\end{table*}

\begin{table*}[]
	\centering
	\begin{tabular}{c|c|c|c|c|c}
		\cline{1-6}
		\centering \textbf{Method}                                                  & \textbf{mAP50 bbox(\%)}    & \textbf{mAP50 segm(\%)}   & \textbf{mAP70 bbox(\%)}   & \textbf{mAP70 segm(\%)}   & \textbf{Train Time}   \\ \hline
		\multicolumn{1}{c|}{Mask R-CNN (output layers)}          		                    & 88.05 $\pm$ 0.32                      & 86.11 $\pm$ 0.29                       & 74.75 $\pm$ 0.19                       & 65.04 $\pm$ 0.62                        & 1h 50m 33s                   \\ \hline
		\multicolumn{1}{c|}{\textbf{Ours Serial}}          		                                    & 88.70 $\pm$ 0.43                      & 87.87 $\pm$ 0.37                       & 71.65 $\pm$ 0.93                       & 64.76 $\pm$ 0.70                        & 37m 18s                   \\ \hline
		\rowcolor[HTML]{79A6D2} 
		\multicolumn{1}{c|}{\textbf{Ours}}          		                                    & 83.63 $\pm$ 1.64                      & 84.50 $\pm$ 1.63                       & 63.33 $\pm$ 1.65                       & 61.54 $\pm$ 0.33                        & 16m 51s                   \\ \hline
	\end{tabular}
	\caption{We report on the performance obtained on HO-3D with \textbf{Ours} and we compare it to \textbf{Mask R-CNN (output layers)} and \textbf{Ours Serial} for the analysis in Sec.~\ref{sec:rpn:indep}.}
	\label{table:rpn_ho3d_2}
\end{table*}

\section{STREAM-BASED INSTANCE SEGMENTATION}
\label{sec:towards_robot}
We now consider a robotic application, in which the robot is tasked to learn new objects on-line, while automatically acquiring training samples. In this case, training data arrive continuously in stream, and the robot is forced to either use them immediately or store them for later use. We investigate to what extent it is possible to reduce the training time and how this affects segmentation performance. 

Because data acquisition takes a considerable amount of time, there is the opportunity to perform, in parallel, some of the processing required for training. In the proposed pipeline, for example, the training protocol \textbf{Ours} has been designed to separate feature extraction and the training of the Kernel-based components. In this case, feature extraction can be performed while images and ground-truth labels are received by the robot. In this section, we investigate to what extent this possibility can be exploited also with the conventional Mask R-CNN architecture.

We compare the proposed \textbf{Ours} with three different Mask R-CNN baselines. Specifically, we consider \textbf{Mask R-CNN (full)} and two variations of \textbf{Mask R-CNN (output layers)} as presented in Sec.~\ref{sec:exp_setup:offline}.

Because images arrive in a stream, similar views of the same objects are represented in subsequent frames. In App.~\ref{sec:towards_robot:no_shuffling} we show that a proper training of \textbf{Mask R-CNN (full)} and \textbf{Mask R-CNN (output layers)} requires that the images are shuffled randomly. This requires storing all images and waiting until the end of the data acquisition process, before starting the training. We hence consider an additional baseline, \textbf{Mask R-CNN (store features)}, in which, similarly to \textbf{Mask R-CNN (output layers)}, we fine-tune the output layers of the RPN and of the detection and segmentation branches. In this case, however, we compute and store the backbone feature maps for each input image during data acquisition to save time. This can be done because, during the fine-tuning, the weights of the backbone remain unaltered. 

Both \textbf{Ours} and \textbf{Mask R-CNN (store features)} can perform the feature extraction while receiving the stream of images: this allows to further reduce the training time. This is possible because the frame rate for feature extraction in both cases is greater than the frame rate of the stream of incoming data. For instance, with \textbf{Ours}, we extract features at $\mathbf{14.7}$ FPS for YCB-Video while the stream of images that is used for training has a frame rate of $3$ FPS (note that the dataset has been collected at $30$ FPS, but we use one image over ten to avoid data redundancy). This allows to completely absorb the time for feature extraction in the time for data acquisition for both approaches. Since the time required for the data acquisition is the same for the two compared methods, we remove it from the training time computation, therefore comparing only the processing time that follows this phase. This represents the time to wait for a model to be ready in the target robotic application. As explained above, the time required for feature extraction cannot be removed in the case of \textbf{Mask R-CNN (full)} and \textbf{Mask R-CNN (output layers)}.

\begin{figure*}
	\centering
  	\includegraphics[width=0.95\linewidth]{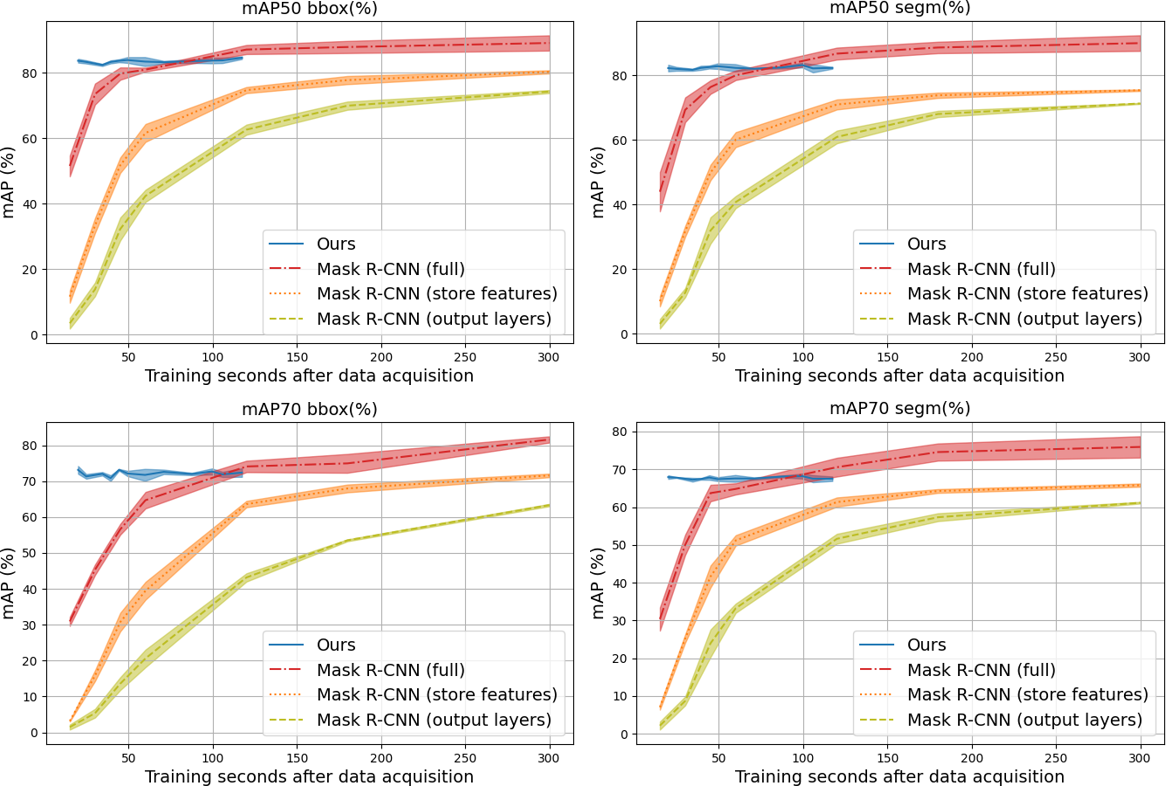}
	\caption{Detection and segmentation \textit{mAPs} for increasing number of \textit{Minibootstrap} iterations for \textbf{Ours} and for increasing training time	of the Mask R-CNN baselines, considering YCB-Video as TARGET-TASK. The plots show the average and the standard deviation of the accuracy obtained over three training sessions with the same parameters.}
	\label{fig:online_vs_mask_ycbv}
\end{figure*}

\begin{figure*}
	\centering
	\includegraphics[width=0.95\linewidth]{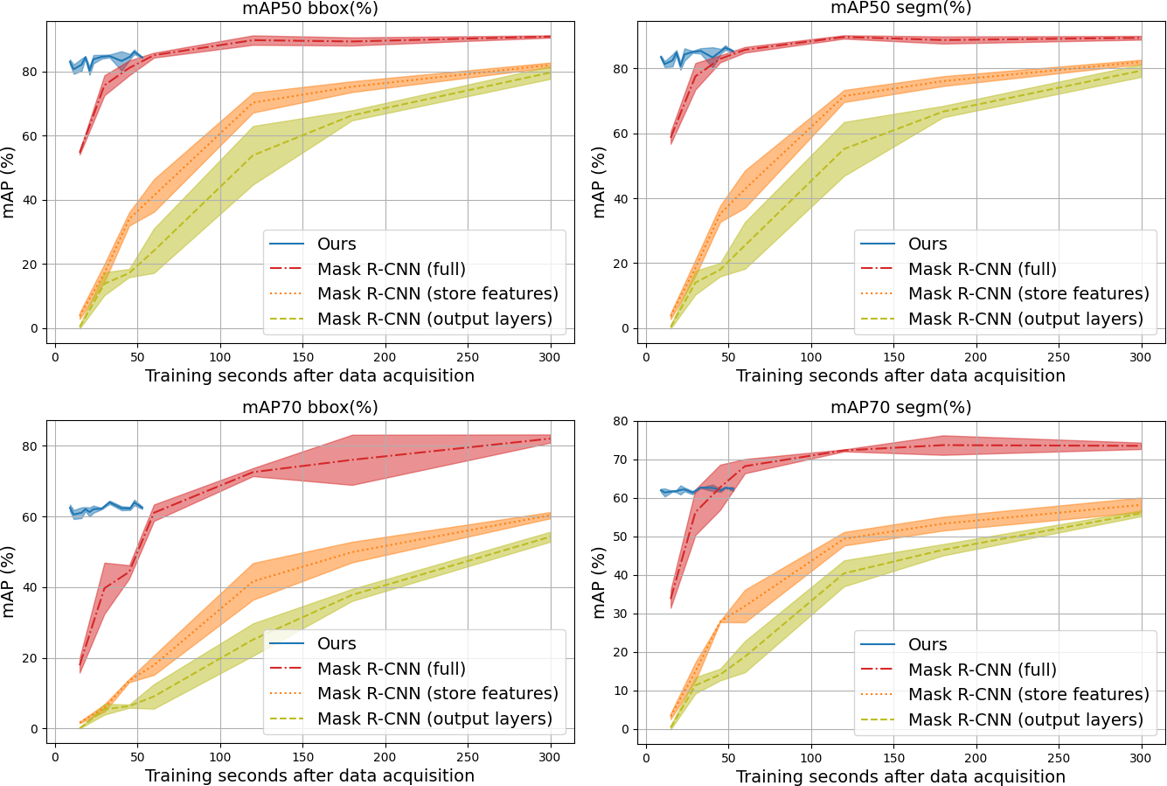}
	\caption{We consider HO-3D as TARGET-TASK and we report the average and the standard deviation of the \textit{mAPs}	over three training sessions with the same parameters for increasing number of \textit{Minibootstrap} iterations for \textbf{Ours}, and for increasing training time of \textbf{Mask R-CNN (full)}, \textbf{Mask R-CNN (output layers)} and \textbf{Mask R-CNN (store features)}.}
	\label{fig:online_vs_mask_ho3d}
\end{figure*}

We present results for this experiment for YCB-Video and HO-3D in Fig.~\ref{fig:online_vs_mask_ycbv} and in Fig.~\ref{fig:online_vs_mask_ho3d}, respectively. Specifically, we compare the performance of the four considered methods for increasing training time. For the Mask R-CNN baselines, we take the accuracy in different moments of the fine-tuning, while, for \textbf{Ours}, we increase the number of iterations of the \textit{Minibootstrap} from $2$ to $15$. In both Fig.~\ref{fig:online_vs_mask_ycbv} and Fig.~\ref{fig:online_vs_mask_ho3d}, we report in the first row the \textit{mAP} trends at \textit{IoU} $50\%$ while in the second row we report the results for \textit{IoU} $70\%$, for both detection and segmentation.

As it can be noticed, \textbf{Ours} achieves the best accuracy for short training time. For instance, in the YCB-Video experiment, if we consider a training time of $\mathbf{{\sim}20s}$, which is the necessary training time if we set the minimum number of \textit{Minibootstrap} iterations $n_B$$=$$2$, \textbf{Ours} achieves a \textbf{mAP} for instance segmentation of \textit{(i)} $\mathbf{{\sim}82.2}$ and \textit{(ii)} $\mathbf{{\sim}67.9}$ for the \textit{IoU} thresholds set to \textit{(i)} $50\%$ and \textit{(ii)} $70\%$. With a similar optimization time, the Mask R-CNN baselines perform quite poorly. For example, \textbf{Mask R-CNN (full)} (which is the best among the baselines) reaches a \textbf{mAP} of \textit{(i)} $\mathbf{{\sim}53.9}$ and \textit{(ii)} $\mathbf{{\sim}39.8}$ for the \textit{IoU} thresholds set to \textit{(i)} $50\%$ and \textit{(ii)} $70\%$. 

Moreover, the plots show that, for all the experiments, \textbf{Mask R-CNN (output layers)} achieves the worst performance, while \textbf{Mask R-CNN (store features)} has a steeper slope. This is due to the fact that this method does not perform the forward pass of the Mask R-CNN backbone for feature extraction. On the contrary, \textbf{Mask R-CNN (full)} presents a better trend than \textbf{Mask R-CNN (output layers)} and \textbf{Mask R-CNN (store features)}. This might be due to the following reasons. Firstly, \textbf{Mask R-CNN (full)} optimizes more parameters of the network. While requiring more time for each training step, this allows to speed-up the optimization process, requiring less iterations on the dataset to achieve comparable accuracy. Secondly, \textbf{Mask R-CNN (full)} performs a warm restart of the the output layers of the RPN, while in the other baselines they are re-initialized from scratch. However, to achieve a similar performance to \textbf{Ours}, \textbf{Mask R-CNN (full)} requires $\mathbf{{\sim}75s}$ for the YCB-Video experiment and $\mathbf{{\sim}50s}$ on HO-3D.

Finally, as it can be noticed, the standard deviations of most of the Mask R-CNN baselines are greater than the ones of \textbf{Ours}. This derives from the fact that while \textbf{Ours} samples features from all the training images, the Mask R-CNN baselines are optimized only on a subset of them due to time constraints (e.g. in the YCB-Video experiment \textbf{Mask R-CNN (full)} processes images at $\mathbf{{\sim}8.0}$ FPS when trained for $\mathbf{1}$ minute). Reducing the number of training images increases the variability of the results.

In the video attached as supplementary material to the manuscript\footnote{\label{video_fn}\url{https://youtu.be/eLatoDWY4OI}}, we show qualitative results to compare \textbf{Ours} to \textbf{Mask R-CNN (full)} when trained for the same time.

\section{ROBOTIC APPLICATION}
\label{sec:robotics_application}
In this section, we describe the pipeline based on the proposed method, that we developed for the iCub~\cite{icub} robot. We set our application in a teacher-learner scenario, in which the robot learns to segment novel objects shown by a human. The proposed application depicts a similar setting to the experiments on HO-3D showing the effectiveness of the approach to learn new objects also in presence of domain shift.

While in the off-line experiments all the input images and the object instances are fixed beforehand, in the application this information is not known in advance. New objects may appear in the scene and, while learning to segment them, the robot has to keep and integrate the knowledge of the classes that are already known. We therefore propose a strategy to process the incoming images and extract corresponding features such that, for each new class, a detection model is trained with \textbf{Ours}, integrating the knowledge of old and new objects. This is done by first training new classifiers on the new classes, considering also the information from the objects already known. Then, the classifiers previously trained on the old classes are updated using features of the new classes.

\begin{figure*}
	\centering
	\includegraphics[width=1.0\linewidth]{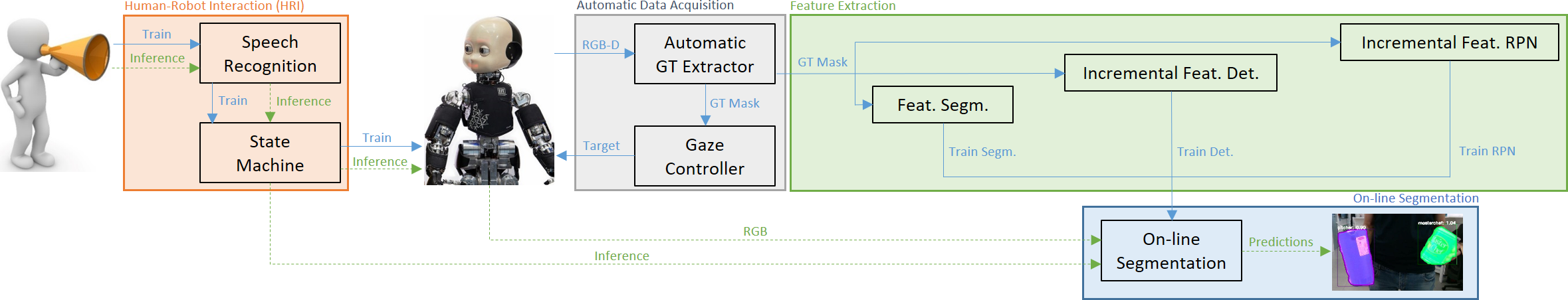}
	\caption{\textbf{Overview of the proposed robotic pipeline} for on-line instance segmentation. At training time (solid arrows), a human teacher shows a new object to the robot, which automatically acquires the ground-truth annotations exploiting the depth information. Then, it extracts the features to train the on-line modules. At inference time (dashed arrows), the robot employs such modules to predict the masks of the images acquired by the camera.}
	\label{fig:pipeline_icub}
\end{figure*}

The proposed application consists of four main modules (the blocks depicted in Fig.~\ref{fig:pipeline_icub}). It allows to train and update an instance segmentation model by: \textit{(i)} automatically collecting ground-truth for instance segmentation with an interactive pipeline for incoming training images, \textit{(ii)} extracting corresponding features and aggregating them such that the information of old and new objects are integrated in the \textit{Minibootstrap} and \textit{(iii)} updating the \textit{On-line RPN}, the \textit{On-line Detection Module} and the \textit{On-line Segmentation Module}. In the next paragraphs, we provide further details for each of the main blocks. 

\noindent
\textbf{Human-Robot Interaction (HRI).} This block allows the human to give commands to the robot with a module for speech recognition (\textit{Speech Recognition} in Fig.~\ref{fig:pipeline_icub}), triggering different states of the system. This allows the user to either teach the robot a new object, by presenting and rotating it in front of the camera (\textit{train}) or to perform \textit{inference}, i.e., to segment objects already known in the scene.

\noindent
\textbf{Automatic Data Acquisition.} When the state of the system is set to \textit{train}, this block extracts a blob of pixels representing the closest object to the robot~\cite{10.3389/frobt.2016.00035}. This is used as ground-truth annotation for the new object that is presented by the human. This blob is computed by exploiting the depth information to segment the object from the background (\textit{Automatic GT Extractor}). Moreover, in order to enhance the background variability in the training images, the extracted blob is also used by the robot to follow the object with the gaze (\textit{Gaze Controller}). To deal with noise in the depth image, we post-process the masks to ensure spatio-temporal coherence between consecutive frames. Specifically, we consider as valid ground-truth masks those overlapping over a certain threshold with the ones of previous and subsequent frames. 

\noindent
\textbf{Feature Extraction.} It is used to extract the features to train the three on-line modules. It relies on the ground-truth masks provided by the \textit{Automatic GT Extractor} and on the corresponding image collected by the robot. This block implements the \textit{Feature Extraction Module} as described in Sec.~\ref{methods:overview}, with some modifications introduced to adapt it to the interactive setting of the demonstration. We describe the major differences in Sec.~\ref{sec:robot:feature}.

\noindent
\textbf{On-line Segmentation.} This block is trained with the proposed approach \textbf{Ours} (see Sec.~\ref{methods:training_protocols}) relying on the features extracted by the \textit{Feature Extraction} block. At inference time, it predicts objects masks on a given image. To do this, similarly to \textbf{Ours}, it relies on Mask R-CNN pre-trained on the MS COCO dataset for feature extraction and on the proposed on-line modules as described in Sec.~\ref{methods:overview}.

\subsection{Incremental Instance Segmentation Learning}
\label{sec:robot:feature}

When a new object has to be learned, the three on-line modules need to be updated. However, only for the \textit{On-line RPN} and the \textit{On-line Detection Module} specific operations are required to integrate the knowledge of the old classes with the new one and to re-train the two modules with the updated information. Instead, for the \textit{On-line Segmentation Module} only the classifier of the novel class must be trained. This is due to the fact that, for each class, the latter extracts masks labels from the ground-truth bounding boxes for that class, while the other modules use all the images in the dataset (see Sec.~\ref{methods:bbox_learning} and Sec.~\ref{methods:oos_learning} for details).

To this end, in the following paragraphs we describe how we adapt the feature extraction procedures for the \textit{On-line RPN} and the \textit{On-line Detection Module} reported in Sec.~\ref{methods:bbox_learning} such that past and novel classes can be properly integrated for the training. We refer the reader to App.~\ref{appendix:prob} for the probabilistic equivalence between the feature sampling procedures of the two on-line modules described in Sec.~\ref{methods:bbox_learning} and the ones presented in this section. Please note that, for each module, we consider training features sampled independently from each training image.

\begin{figure*}
	\centering
	\includegraphics[width=1.0\linewidth]{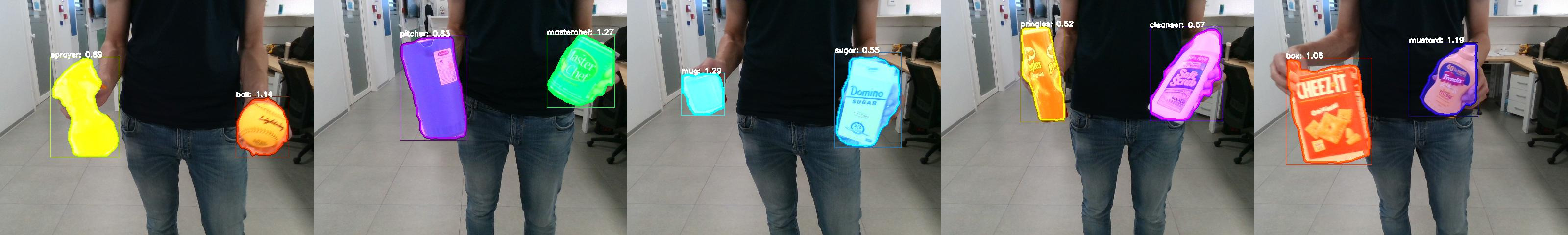}
	\caption{Predictions on test images from the incremental application deployed on the iCub.}
	\label{fig:inference_icub}
\end{figure*}

\begin{figure*}
	\centering
	\includegraphics[width=0.8\linewidth]{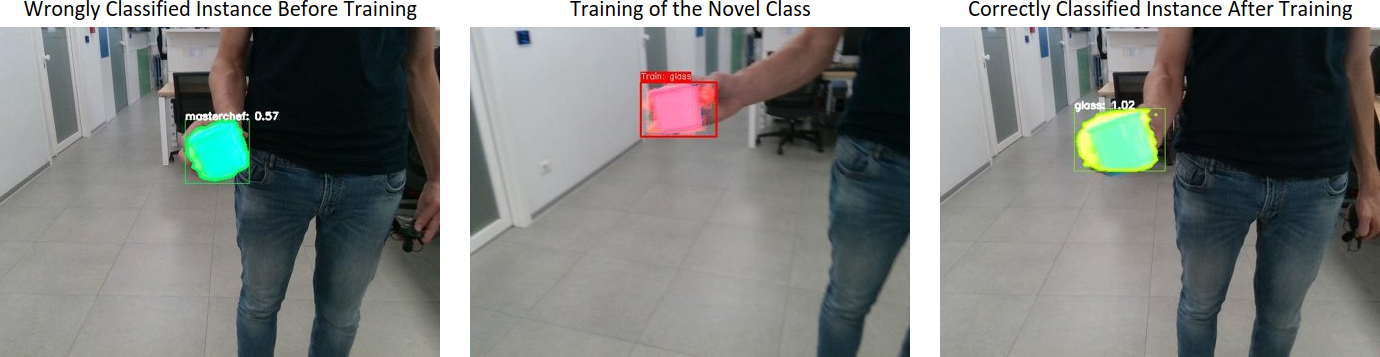}
	\caption{\textbf{Dealing with False Positives}. \textit{Left image}: an unknown object (a \textit{glass}) is misclassified (as a \textit{masterchef}). \textit{Center}: training. The robot is provided with the correct label and a demonstration of the object. \textit{Right}: after training the new object is correctly classified.}
	\label{fig:inference_fp}
\end{figure*}

For the sake of simplicity, in the following analysis, we consider a single incremental task scenario where a sequence of images representing an instance of a new class is shown to the robot, which is required to learn the new object at the end of the demonstration. Specifically, the robot has already been trained on $N-1$ classes and must learn the $N^{th}$ object. However, Alg.~\ref{alg:inc_rpn} and Alg.~\ref{alg:inc_det} describe in detail the general procedures, which are suitable also if multiple objects must be learned simultaneously. Once the training features for these two modules have been computed with the described approaches, they can be trained with the same procedure described in Sec.~\ref{methods:bbox_learning}, namely with the steps 2 and 3 of the \textit{Minibootstrap} procedure (see App.~\ref{appendix:minibootstrap}).

Positive samples for the $N^{th}$ object, as defined in Sec.~\ref{methods:bbox_learning} for FALKON classifiers and RLS regressors in the \textit{On-line RPN} and in the \textit{On-line Detection Module}, are taken from the current image stream and are not affected by previous sequences.
Therefore, we stick to the method described in Sec.~\ref{methods:bbox_learning} for their extraction. Instead, for the computation of the negative features we design two different procedures that we describe in the following paragraphs.

\noindent
\textbf{On-line RPN.} When learning the $N^{th}$ class, we first collect features for \textit{On-line RPN} training for that class. We do this by extracting convolutional features from the images of the associated sequence and sub-sample them as described in Sec.~\ref{methods:bbox_learning}. Then, we need to integrate the extracted features with those from the previous $N-1$ classes for each anchor, such that \textit{(i)} the number and size of \textit{Minibootstrap} batches are kept fixed and \textit{(ii)} the number of negative samples per-image is kept balanced. To this end, we randomly remove a fraction of the samples collected for the \textit{Minibootstrap} batches of the previous $N-1$ classes and substitute them with those for the $N^{th}$ class. More details about this procedure can be found in App.~\ref{appendix:incremental_rpn}.

\noindent
\textbf{On-line Detection.} Similarly to what is done for the \textit{On-line RPN}, when the $N^{th}$ class arrives, we extract convolutional features from the images of the associated sequence and sub-sample them as described in Sec.~\ref{methods:bbox_learning}. Then, the extracted features have to be integrated with those from the previous $N-1$ classes. This is done in a twofold way: \textit{(i)} we create the $N^{th}$ dataset to train a classifier for the new object integrating the extracted features with a subset of the previous $N-1$ ones and \textit{(ii)} we update the $N-1$ datasets for training the previous classes with features from the $N^{th}$. As for the \textit{On-line RPN}, the number and size of \textit{Minibootstrap} batches are kept fixed and we balance the number of per-image negative samples. More details about this procedure can be found in App.~\ref{appendix:incremental_ood}.

\subsection{Discussion and Qualitative Results}
\label{sec:robot:discussion}
We design the incremental feature extraction procedures for the \textit{On-line RPN} and the \textit{On-line Detection Module} to be analogous to the ones used in the off-line experiments (batch procedures), such that, training the on-line modules with \textit{Minibootstrap} batches obtained with the former, provides comparable models to the ones obtained with the batch procedures (and therefore, comparable accuracy). This is due to the fact that a set of negative samples has the same probability to end up in the \textit{Minibootstrap} batches of a classifier (either of the \textit{On-line RPN} or of the \textit{On-line Detection Module}), either using the batch pipelines as described in Sec.~\ref{methods:bbox_learning} or the incremental procedures presented in the previous section. This is demonstrated in App.~\ref{appendix:prob}. Specifically, we demonstrate that the per-image negative selection probabilities with the two procedures are equivalent, because each image is considered independently in both cases. This proves their equivalence.

We qualitatively show the effectiveness of the incremental pipeline by deploying it on the iCub robot. We train ten object instances and we report the results of the inference on some exemplar frames in Fig.~\ref{fig:inference_icub}. A video of the complete demonstration, comprising the training of all the considered objects and the inference of the trained models, is attached as supplementary material to the manuscript\footref{video_fn}. In Fig.~\ref{fig:inference_fp}, we show how the proposed incremental approach allows to deal with false positive predictions. Key to achieve this is the re-training of the $N-1$ classifiers for previous classes when the $N^{th}$ object arrives. Indeed, integrating data from the $N^{th}$ object when updating the previous $N-1$ allows to strongly reduce the amount of false predictions at inference time.

\section{CONCLUSIONS}
\label{sec:conclusions}
The ability of rapidly adapting their visual system to novel tasks is an important requirement for robots operating in dynamic environments. While state-of-the art approaches for visual tasks mainly focus on boosting performance, a relatively small amount of methods are designed to reduce training time. In this perspective, we presented a novel pipeline for fast training of the instance segmentation task. The proposed approach allows to quickly learn to segment novel objects also in presence of domain shifts. We designed a two-stage hybrid pipeline to operate in the typical robotic scenario where streams of data are acquired by the camera of the robot. Indeed, our pipeline allows to shorten the total training time by extracting a set of convolutional features during the data acquisition and to use them in a second step to rapidly train a set of Kernel-based classifiers. 

We benchmarked our results on two robotics datasets, namely YCB-Video and HO-3D. On these datasets, we provided an extensive empirical evaluation of the proposed approach to evaluate different training time/accuracy trade-offs, comparing results against previous work~\cite{ceola2020segm} and several Mask R-CNN baselines.

Finally, we demonstrated the application of this work on a real humanoid robot. At this aim, we adapted the fast training pipeline for incremental region proposal adaptation and instance segmentation, showing that the robot is able to learn new objects following a short interactive training session with a human teacher.

\appendices

\section{}
\label{appendix:summary_tab}

\begin{table*}[h]
	\centering
	\begin{tabular}{c|m{11.5cm}|c}
		\cline{1-3}
		\centering \textbf{Method}                                                  & \centering \textbf{Training Protocol}    &  \textbf{Section}   \\ \hline
		\textbf{Ours}          		                                    &
        This is the proposed approach. It is composed of two steps. One for \textit{feature extraction} and one for simultaneous \textit{on-line training} of the proposed methods for region proposal, object detection and mask prediction.
        & \ref{methods:oos_learning}\\ \hline
		\textbf{Ours Serial}          		                                & 
		It is composed of the same modules as \textbf{Ours}, but it differs on the training protocol. This is composed of two steps for \textit{feature extraction}, one for the \textit{on-line training} of the proposed method for region proposal and one to \textit{train} the modules for on-line detection and segmentation.
		& \ref{sec:rpn:indep} \\ \hline
		\textbf{O-OS}                          		                    & 
		This is the approach proposed in \cite{ceola2020segm}. It is composed of two steps. One for \textit{feature extraction} and one for \textit{on-line training} of the modules for object detection and mask prediction.& \ref{sec:rpn:gain}\\ \hline
		\textbf{Mask R-CNN (full)}          		                                & This protocol relies on Mask R-CNN pre-trained on the FEATURE-TASK as a warm-restart to train Mask R-CNN on the TARGET-TASK. & \ref{sec:exp_setup:offline} \\ \hline
		\textbf{Mask R-CNN (output layers)}          		                        & Starting from the Mask R-CNN weights pre-trained on the FEATURE-TASK, it fine-tunes the output layers of the RPN and of the detection and segmentation branches on the TARGET-TASK. & \ref{sec:exp_setup:offline} \\ \hline
		\textbf{Mask R-CNN (store features)}          		        & Similarly to \textbf{Mask R-CNN (output layers)}, it fine-tunes the output layers of the RPN and of the detection and segmentation branches. However, since the weights of the backbone remain unaltered, it computes and stores the backbone feature maps for each input image during data acquisition. & \ref{sec:towards_robot} \\ \hline
	\end{tabular}
	\caption[]{Training protocols overview.}
	\label{table:summary_architecture}
\end{table*}

In Tab.~\ref{table:summary_architecture} and in Tab.~\ref{table:metric}, we overview the training protocols and the acronyms for \textit{mAP} evaluation used in this work.

\newcolumntype{M}[1]{>{\centering\arraybackslash}m{#1}}

\begin{table*}[]
	\centering
	\begin{tabular}{M{3.6cm}|c|c}
		\cline{1-3}
		\textbf{\textit{Intersection over Union (IoU)}} 
		\textbf{with ground-truth} & \textbf{Object Detection}    & \textbf{Instance Segmentation}   \\ \hline
		$\mathbf{50\%}$ & mAP50 bbox(\%)         		                                & mAP50 segm(\%) \\ \hline
		$\mathbf{70\%}$ &mAP70 bbox(\%)                        		                    & mAP70 segm(\%) \\ \hline
	\end{tabular}
	\caption[]{Object detection and segmentation metrics taxonomy.}
	\label{table:metric}
\end{table*}

\section{}

\label{appendix:ho3d}
In this appendix, we report the sequences considered in the experiments on the HO-3D dataset.

\begin{itemize}
    \item \textbf{Training} sequences: \textit{ABF10}, \textit{ABF11}, \textit{ABF12}, \textit{ABF13}, \textit{BB10}, \textit{BB11}, \textit{BB12}, \textit{BB13}, \textit{GPMF10}, \textit{GPMF11}, \textit{GPMF12}, \textit{GPMF13}, \textit{GSF10}, \textit{GSF11}, \textit{GSF12}, \textit{GSF13}, \textit{MC1}, \textit{MC2}, \textit{MC4}, \textit{MC5}, \textit{MDF10}, \textit{MDF11}, \textit{MDF12}, \textit{MDF13}, \textit{ShSu10}, \textit{ShSu12}, \textit{ShSu13}, \textit{ShSu14}, \textit{SM2}, \textit{SM3}, \textit{SM4}, \textit{SMu1}, \textit{SMu40}, \textit{SMu41}.
    \item \textbf{Validation} sequences: \textit{ABF13}, \textit{BB13}, \textit{GPMF13}, \textit{GSF13}, \textit{MC5}, \textit{MDF13}, \textit{ShSu14}, \textit{SM4}, \textit{SMu41}.
    \item \textbf{Test} sequences: \textit{ABF14}, \textit{BB14}, \textit{GPMF14}, \textit{GSF14}, \textit{MC6}, \textit{MDF14}, \textit{SiS1}, \textit{SM5}, \textit{SMu42}.
\end{itemize}


\section{}
\label{sec:towards_robot:no_shuffling}
\begin{table*}[]
	\centering
	\begin{tabular}{c|c|c|c|c|c|c}
		\cline{1-7}
		\multicolumn{2}{c|}{\textbf{Method}}                                                                  & \textbf{mAP50 bbox(\%)}    & \textbf{mAP50 segm(\%)}      & \textbf{mAP70 bbox(\%)}       & \textbf{mAP70 segm(\%)}       & \textbf{Train Time}   \\ \hline
		\multirow{2}*{1 Epoch} &\multicolumn{1}{c|}{Mask R-CNN (full)}          		                            & 89.21 $\pm$ 0.77                      & 90.75 $\pm$ 0.80                       & 83.27 $\pm$ 0.54                       & 78.83 $\pm$ 1.70                        & 35m 8s                   \\ 
		 &\multicolumn{1}{c|}{Mask R-CNN (output layers)}          		                            & 82.78 $\pm$ 0.50                      & 79.38 $\pm$ 0.40                       & 75.04 $\pm$ 0.97                       & 68.42 $\pm$ 0.55                        & 26m 48s                   \\ \hline
		\multirow{2}*{\shortstack{1 Epoch \\ No Shuffling}}  &\multicolumn{1}{c|}{Mask R-CNN (full)}          		                            & 46.29 $\pm$ 1.26                      & 43.93 $\pm$ 0.84                       & 28.66 $\pm$ 1.89                       & 32.83 $\pm$ 1.37                        & 33m 41s                   \\ 
		&\multicolumn{1}{c|}{Mask R-CNN (output layers)}          		                            & 69.63 $\pm$ 0.48                      & 66.03 $\pm$ 0.68                       & 38.48 $\pm$ 2.88                       & 54.70 $\pm$ 0.46                        & 26m 39s                   \\ \hline
		\rowcolor[HTML]{79A6D2} 
		\multicolumn{2}{c|}{\textbf{Ours}}          		                                    & 83.66 $\pm$ 0.84                      & 83.06 $\pm$ 0.92                       & 72.97 $\pm$ 1.02                       & 68.11 $\pm$ 0.29                          & 13m 53s                      \\ \hline
	\end{tabular}
	\caption{Comparison between the training of \textbf{Ours} and the Mask R-CNN baselines trained for one epoch and for one epoch without shuffling the input images, considering YCB-Video as TARGET-TASK.}
	\label{table:towards_robot:no_shuffling_ycbv}
\end{table*}

\begin{table*}[]
	\centering
	\begin{tabular}{c|c|c|c|c|c|c}
		\cline{1-7}
		\multicolumn{2}{c|}{\textbf{Method}}                                                                  & \textbf{mAP50 bbox(\%)}    & \textbf{mAP50 segm(\%)}      & \textbf{mAP70 bbox(\%)}       & \textbf{mAP70 segm(\%)}       & \textbf{Train Time}   \\ \hline
		\multirow{2}*{1 Epoch} &\multicolumn{1}{c|}{Mask R-CNN (full)}          		                            & 92.21 $\pm$ 0.88                      & 90.70 $\pm$ 0.17                       & 86.73 $\pm$ 0.71                       & 77.25 $\pm$ 0.62                        & 38m 38s                   \\
		&\multicolumn{1}{c|}{Mask R-CNN (output layers)}          		                            & 86.77 $\pm$ 0.68                      & 85.45 $\pm$ 0.64                       & 70.57 $\pm$ 0.41                       & 63.57 $\pm$ 0.43                        & 17m 53s                   \\ \hline
		\multirow{2}*{\shortstack{1 Epoch \\ No Shuffling}}  &\multicolumn{1}{c|}{Mask R-CNN (full)}          		                            & 6.72 $\pm$ 2.35                      & 6.53 $\pm$ 2.33                       & 6.24 $\pm$ 2.21                       & 6.01 $\pm$ 2.18                        & 37m 30s                   \\
		&\multicolumn{1}{c|}{Mask R-CNN (output layers)}          		                            & 19.28 $\pm$ 2.33                      & 21.01 $\pm$ 0.75                       & 11.03 $\pm$ 1.70                       & 16.74 $\pm$ 1.26                        & 17m 52s                   \\ \hline
		\rowcolor[HTML]{79A6D2} 
		\multicolumn{2}{c|}{\textbf{Ours}}          		                                    & 83.63 $\pm$ 1.64                      & 84.50 $\pm$ 1.63                       & 63.33 $\pm$ 1.65                       & 61.54 $\pm$ 0.33                        & 16m 51s                   \\ \hline 
	\end{tabular}
	\caption{Comparison between the training of \textbf{Ours} and the Mask R-CNN baselines trained for one epoch and for one epoch without shuffling the input images, considering  HO-3D as TARGET-TASK.}
	\label{table:towards_robot:no_shuffling_ho3d}
\end{table*}

In the stream-based scenario, data is used as soon as it is received (Sec.~\ref{sec:towards_robot}). In this case, the Mask R-CNN baselines can be trained only for one epoch and their weights can be updated only when a new image is received. Therefore, we compare the performance achieved by \textbf{Ours} against the Mask R-CNN baselines, training \textbf{Mask R-CNN (output layers)} and \textbf{Mask R-CNN (full)} for one epoch and without shuffling the input images.

Result are reported in Tab.~\ref{table:towards_robot:no_shuffling_ycbv} for YCB-Video and in Tab.~\ref{table:towards_robot:no_shuffling_ho3d} for HO-3D. As it can be noticed, while training the Mask R-CNN baselines for just one epoch allows to achieve a performance similar to the one provided in the benchmarks (Sec.~\ref{sec:results}), shuffling the input images turns out to be crucial. Indeed, in both cases, the accuracy provided by the Mask R-CNN baselines drops when the input images are not shuffled, while \textbf{Ours} is not affected by this constraint. Moreover, shuffling is particularly critical for the Mask R-CNN baselines on the HO-3D dataset because training objects are shown subsequently, one by one, as in the target teacher-learner setting (Sec.~\ref{sec:robotics_application}). Therefore, in the stream-based scenario, such baselines cannot be trained in practice.

\section{}
\label{appendix:prob}
In this appendix, we prove the probabilistic equivalence of the two sampling procedures which are used in the \textit{Minibootstrap} and in the incremental feature extraction pipelines for both the \textit{On-line RPN} and for the \textit{On-line Detection Module}.

We consider a pool of tensors $S_0$ and a set of tensors $\hat{S} \subseteq S_0$. We compute the probability of sampling $\hat{S}$ from $S_0$ with the two following procedures:
\begin{itemize}
    \item We sample $|\hat{S}|$ tensors from $S_0$. We refer to the probability that the sampled tensors are equal to $\hat{S}$ as $P(\hat{S} \sim S_0)$.
    \item  We recursively sample $m$ sets from $S_0$ and we obtain the pools $S_1,..,S_m$ such that $|S_0| \geq |S_1| \geq ... \geq |S_m| \geq |\hat{S}|$. Namely, for each $S_i$, with $i$ in $0, ..., m-1$, $S_{i+1}$ is a random sample of size $|S_{i+1}|$ of $S_{i}$. Finally, we sample $|\hat{S}|$ tensors from $S_m$. We refer to the probability that the sampled tensors are equal to $\hat{S}$ as $P(\hat{S} \sim S_m)$.
\end{itemize}

We note that, for the \textit{On-line RPN} and for the \textit{On-line Detection Module}, the pool of tensors $S_0$ represents the whole set of features associated to an image. $S_1,..,S_m$, instead, represent consecutive sub-samples of $S_0$ in the incremental feature extraction pipelines. Finally, $\hat{S}$ correspond to the final per-image set of features chosen for training the on-line modules either with the \textit{Minibootstrap} (as presented in~\ref{methods:bbox_learning}) or with the incremental pipelines.

We prove that $P(\hat{S} \sim S_0)$ is equal to $P(\hat{S} \sim S_m)$.
\begin{proof}
$P(\hat{S} \sim S_0)$ can be computed as follows:
\begin{equation}
    P(\hat{S} \sim S_0) = \frac{1}{\binom{|S_0|}{|\hat{S}|}}
\end{equation}
Instead, due to the law of total probability, we can decompose $P(\hat{S} \sim S_m)$ as follows (note that if $\hat{S}$ is not a subset of $S_{m}$, $P(\hat{S} \sim S_m | \hat{S} \not \subseteq S_{m}) = 0$):
\begin{equation}
    \label{eq:prob}
    P(\hat{S} \sim S_m) = P(\hat{S} \sim S_m | \hat{S} \subseteq S_{m}) \times P ( \hat{S} \subseteq S_{m})\\
\end{equation}
Again, due to the law of total probability, we can decompose $P(\hat{S}\subseteq S_{m})$ from equation~\ref{eq:prob} as follows (note that, for each $i$ in $0, ..., m-1$, $P(\hat{S} \subseteq S_{i} | \hat{S} \not \subseteq S_{i-1}) = 0$):
\begin{equation}
    \begin{split}
     P(\hat{S}\subseteq S_{m}) &= P ( \hat{S} \subseteq S_{m} | \hat{S} \subseteq S_{m-1}) \times P ( \hat{S} \subseteq S_{m-1})\\
                &=\prod\limits_{k=m}^{1}  P(\hat{S} \subseteq S_k | \hat{S} \subseteq S_{k-1}) \times P (\hat{S} \subseteq S_{0})
    \end{split}
\end{equation}
Note that $P (\hat{S} \subseteq S_{0}) = 1$ by definition (i.e., $\hat{S}$ is always in $S_0$). Therefore:
\begin{equation}
    \begin{split}
    P(\hat{S}\subseteq S_{m}) &=\prod\limits_{k=m}^{1}  P(\hat{S} \subseteq S_k | \hat{S} \subseteq S_{k-1})\\
    \end{split}
\end{equation}
We note that $\binom{|S_{k-1}| - |\hat{S}|}{|S_k| - |\hat{S}|}$ is the total number of feasible samples s.t. $\hat{S} \subseteq S_k$ given that $\hat{S} \subseteq S_{k-1}$. Namely, we fix $\hat{S}$ in $S_k$ and we compute the number of possible combinations of the remaining $|S_k| - |\hat{S}|$ tensors that can be in $S_k$ sampled from the remaining pool of size $|S_{k-1}| - |\hat{S}|$.
Therefore:
\begin{equation}
    \begin{split}
    P(\hat{S} \subseteq S_k | \hat{S} \subseteq S_{k-1}) = \frac{\binom{|S_{k-1}| - |\hat{S}|}{|S_k| - |\hat{S}|}}{\binom{|S_{k-1}|}{|S_k|}}\\
    \end{split}
\end{equation}
We can decompose the component in the product as follows:
\begin{equation}
    \begin{split}
    \frac{\binom{|S_{k-1}| - |\hat{S}|}{|S_k| - |\hat{S}|}}{\binom{|S_{k-1}|}{|S_k|}}= \frac{(|S_{k-1}|-|\hat{S}|)!}{(|S_{k}|-|\hat{S}|)!} \times \frac{|S_{k}|!}{|S_{k-1}|!}
    \end{split}
\end{equation}
Note that, for each of these elements with $k \neq 1$ and $k \neq m$, if we multiply it by the element at $k-1$ and by the element at $k+1$, all the elements are simplified. Therefore, we can rewrite $P(\hat{S} \sim S_m)$ from equation~\ref{eq:prob} as:
\begin{equation}
    \begin{split}
    P(\hat{S} \sim S_m&) =P(\hat{S} \sim S_m | \hat{S} \subseteq S_{m}) \times P ( \hat{S} \subseteq S_{m})\\
    &= \frac{1}{\binom{|S_m|}{|\hat{S}|}} \times \prod\limits_{k=m}^{1}  \frac{\binom{|S_{k-1}| - |\hat{S}|}{|S_k| - |\hat{S}|}}{\binom{|S_{k-1}|}{|S_k|}}\\
    &= \frac{(|S_m|-|\hat{S}|)! \times |\hat{S}|!}{|S_m|!} \times \frac{|S_m|! \times (|S_{0}|-|\hat{S}|)!}{(|S_m|-|\hat{S}|)! \times S_{0}!}\\
    &= \frac{|\hat{S}|! \times (|S_{0}|-|\hat{S}|)!}{|S_{0}|!}\\
    &= \frac{1}{\binom{|S_{0}|}{|\hat{S}|}}
    \end{split}
\end{equation}
This concludes our proof, since:
\begin{equation}
    P(\hat{S} \sim S_0) = P(\hat{S} \sim S_m)
\end{equation}
\end{proof}

\section{}
\label{appendix:minibootstrap}

In Alg.~\ref{alg:minib}, we report the pseudo-code of the \textit{Minibootstrap}~\cite{maiettini2019a} procedure. Note that, we use the Sample(\textit{Set, Sample size}) function to extract \textit{Sample size} random tensors from the given \textit{Set}. We will use this function also in Alg.~\ref{alg:inc_rpn} and Alg.~\ref{alg:inc_det}.

\begin{algorithm*}
\caption{\textbf{Minibootstrap} Pseudo-code for the \textit{Minibootstrap} in the off-line experiments. See~\cite{maiettini2019a} for further details.}
\label{alg:minib}
\begin{algorithmic}
\renewcommand{\algorithmicrequire}{\textbf{Input}:}
\renewcommand{\algorithmicensure}{\textbf{Output}:}
\Require $BS$, $n_B$: size and number of \textit{Minibootstrap} batches\\
        $I$: set of training images\\
        $N$: number of classes
\Ensure  $M$: $N$ trained classifiers
\State \textit{\textbf{Stage 1}: Feature extraction}
\For {$n = 1$ to $N$} \Comment{\textit{Initialize $N$ empty sets of per class \textbf{positives} and \textbf{negatives} features}}
    \State $Pos[n] \gets \emptyset$; $Neg[n] \gets \emptyset$
\EndFor
\For {$i = 1$ to $|I|$}
    \For {$n = 1$ to $N$}
        \If {($I[i]$ has positives of class $n$ $Pos_{i,n}$)}
            \State $Pos[n] \gets Pos[n] \bigcup Pos_{i,n}$ \Comment{\textit{\textbf{Add positives} for class $n$ from the $i^{th}$ image}}
        \EndIf 
        \State $Neg[n] \gets Neg[n] \bigcup$ Sample$(Neg_{i,n}, \lceil \frac{n_B{\times}BS}{|I|} \rceil)$         \Comment{\textit{\textbf{Add negatives} for class $n$ from the $i^{th}$ image}}
    \EndFor
\EndFor
\State \textit{\textbf{Stage 2}: Shuffling and batches creation}
\For {$n = 1$ to $N$}
    \State $Neg[n] \gets$ Sample$(Neg[n], n_B{\times}BS)$
    \State $Neg[n] \gets$ Split$(Neg[n], n_B, BS)$ \Comment{\textit{\textbf{Split} $Neg[n]$ in $n_B$ batches of size $BS$}}
\EndFor
\State \textit{\textbf{Stage 3}: Classifiers training}
\For {$n = 1$ to $N$}
    \State $F[n] \gets Pos[n] \bigcup Neg[n][1]$
    \State $M[n] \gets $TrainClassifier$(F[n])$     \Comment{\textit{\textbf{Train} classifier using the first batch}}
    \State $Neg_{chosen}[n] \gets$ PruneEasy$(M[n], Neg[n][1])$ \Comment{\textit{\textbf{Prune easy negatives} from $Neg[n][1]$ using $M[n]$}}
    \For {$j = 2$ to $n_B$}
        \State $N^H \gets$ SelectHard$(M[n], Neg[n][j])$ \Comment{\textit{\textbf{Select hard negatives} from $Neg[n][j]$ using $M[n]$}}
        \State $F[n] \gets Pos[n] \bigcup Neg_{chosen}[n] \bigcup N^H$\Comment{\textit{\textbf{Add hard negatives} from $Neg[n][j]$ to the training set}}
        \State $M[n] \gets $TrainClassifier$(F[n])$         \Comment{\textit{\textbf{Train} classifier using the new dataset}}
        \State $Neg_{chosen}[n] \gets Neg_{chosen}[n] \bigcup N^H$ \Comment{\textit{\textbf{Update} the chosen \textbf{negatives}}}
        \State $Neg_{chosen}[n] \gets$ PruneEasy$(M[n], Neg_{chosen}[n])$ \Comment{\textit{\textbf{Prune easy negatives} from $Neg_{chosen}[n]$ using $M[n]$}}
    \EndFor
\EndFor
\State \textit{Return $M$} \Comment{\textit{\textbf{Return} the final classifiers}}
\end{algorithmic}
\end{algorithm*}

\section{}
\label{appendix:incremental_rpn}

We report the pseudo-code for the incremental feature extraction pipeline for the \textit{On-line RPN}. Since the procedure is equal for all the considered anchors, in Alg.~\ref{alg:inc_rpn} we report the algorithm for a generic anchor $a$.

\begin{algorithm*}
\caption{\textbf{Incremental On-line RPN} Pseudo-code of the \textit{incremental} feature extraction procedure for the \textit{On-line RPN}.}
\label{alg:inc_rpn}
\begin{algorithmic}
\renewcommand{\algorithmicrequire}{\textbf{Input}:}
\renewcommand{\algorithmicensure}{\textbf{Output}:}
\Require  $BS$, $n_B$: size and number of \textit{Minibootstrap} batches\\
        $Pos^{t-1}$, $Neg^{t-1}$: \textbf{positive} and per-image \textbf{negative} features at iteration $t-1$ for a generic anchor $a$\\
        $I^t$: $t^{th}$ sequence of training images\\
        $\#IMG^t = \#IMG^{t-1} + |I^t|$: total number of training images seen in the previous iterations and in this sequence
\Ensure  $Pos^t$, $Neg^t$: \textbf{positive} and per-image \textbf{negative} features at iteration $t$ for anchor $a$
\State \textit{\textbf{Stage 1}: Sample negative features from the ones at iteration $t-1$}
\For {$j = 1$ to $\#IMG^{t-1}$}
    \State $Neg^t[j] \gets$ Sample$(Neg^{t-1}[j], \lceil \frac{n_B{\times}BS}{\#IMG^t} \rceil)$
\EndFor
\State \textit{\textbf{Stage 2}: Append new features from sequence $I^t$}
\State \textit{$Pos^t \gets Pos^{t-1}$}
\For {$i = 1$ to $|I^t|$}
    \If {$I^t[i]$ has positives for anchor $a$ $Pos_{I^t[i]}$}
        \State $Pos^t \gets Pos^t \bigcup Pos_{I^t[i]}$ \Comment{\textit{\textbf{Add positives} for anchor $a$ from the image $I^t[i]$}}
    \EndIf 
    \State $Neg^t \gets Neg^t \bigcup$ Sample$(Neg_{I^t[i]}, \lceil \frac{n_B{\times}BS}{\#IMG^t} \rceil)$ \Comment{\textit{\textbf{Add negatives} for anchor $a$ from the image $I^t[i]$}}
\EndFor
\State \textit{Return $Pos^t, Neg^t$}   \Comment{\textit{\textbf{Return} positive and negative features at iteration $t$}}
\end{algorithmic} 
\end{algorithm*}

\section{}
\label{appendix:incremental_ood}

In Alg.~\ref{alg:inc_det}, we report the pseudo-code for the incremental feature extraction pipeline for the \textit{On-line Detection Module}.

\begin{algorithm*}
\caption{\textbf{Incremental On-line Detection} \textit{Incremental} feature extraction pseudo-code for the \textit{On-line Detection Module}.}
\label{alg:inc_det}
\begin{algorithmic}
\renewcommand{\algorithmicrequire}{\textbf{Input}:}
\renewcommand{\algorithmicensure}{\textbf{Output}:}
\Require $BS$, $n_B$: size and number of the \textit{Minibootstrap} batches\\
        $N^t$: number of classes. It comprises the $N^{t-1}$ classes known at iteration $t-1$ and the novel classes at iteration $t$\\
        $Pos^{t-1}$, $Img_{Neg}^{t-1}$: sets of $N^{t-1}$ (one for each class) \textbf{positive} and per-image \textbf{negative} features at iteration $t-1$\\
        $Img_{Buf}^{t-1}$: set of per-image \textbf{buffer} features at iteration $t-1$\\
        $I^t$: $t^{th}$ sequence of training images\\
        $\#IMG^t = \#IMG^{t-1} + |I^t|$: total number of training images in the previous iterations and in this sequence
\Ensure $Pos^t$, $Neg^t$: $N^t$ sets of \textbf{positive} and \textbf{negative} training features at iteration $t$
\State \textit{\textbf{Stage 1}: Sample negative per-image features from the previous sequences}
\For {$i = 1$ to $\#IMG^{t-1}$}
    \For {$n = 1$ to $N^t$}
        \If {$n \leq N^{t-1}$}
            \State $Img_{Neg}^{t}[n][i] \gets$ Sample$(Img_{Neg}^{t-1}[n][i], \lceil \frac{n_B{\times}BS}{\#IMG^t} \rceil)$
        \Else
            \State $Img_{Neg}^{t}[n][i] \gets \emptyset$ \Comment{\textit{Set per-image negatives of the old sequences to an \textbf{empty set} for the \textbf{new classes}}}
        \EndIf
    \EndFor
    \State $Img_{Buf}^t[i] \gets$ Sample$(Img_{Buf}^{t-1}[i], \lceil \frac{n_B{\times}BS}{\#IMG^t} \rceil)$
\EndFor
\State \textit{$Pos^t \gets Pos^{t-1} \bigcup \bigcup_{i=0}^{N^{t}-N^{t-1}} \emptyset$} \Comment{\textit{\textbf{Compute $Pos^t$ from $Pos^{t-1}$} adding an \textbf{empty set} for each \textbf{new class}}}
\State \textit{\textbf{Stage 2}: Sample features from the $t^{th}$ sequence of images}
\For {$i = 1$ to $|I^t|$}
    \For {$n = 1$ to $N^t$}
        \If {$I^t[i]$ has positives of class $n$ $Pos_{I^t[i],n}$} 
            \State $Pos^t[n] \gets Pos^t[n] \bigcup Pos_{I^t[i],n}$ \Comment{ \textit{\textbf{Add positives} for class $n$ from $I^t[i]$}}
            \State $Img_{Neg}^t[n] \gets Img_{Neg}^t[n] \bigcup$ Sample$(Neg_{I^t[i],n}, \lceil  \frac{n_B{\times}BS}{\#IMG^t} \rceil)$ \Comment{\textit{\textbf{Add} per-image \textbf{negatives} for class $n$}}
        \Else
            \State $Img_{Neg}^t[n] \gets Img_{Neg}^t[n] \bigcup \emptyset$
        \EndIf
    \EndFor
    \State $Img_{Buf}^t \gets Img_{Buf}^t \bigcup$ Sample$(All\_feat_{I^t[i]}, \lceil \frac{n_B{\times}BS}{\#IMG^t} \rceil)$  \Comment{\textit{\textbf{Add} per-image \textbf{buffer negatives} from $I^t[i]$}}
\EndFor
\State \textit{\textbf{Stage 3}: Fill \textbf{negative} batches}
\For {$n = 1$ to $N^t$}
    \State{$Neg^t[n] \gets \emptyset$}
    \For {$i = 1$ to $\#IMG^t$}
        \If {($Img_{Neg}[n][i] \neq \emptyset $)}    \Comment{\textit{If any, add to the \textbf{negatives} of class $n$, per-image \textbf{negatives} for class $n$}}
            \State $Neg^t[n] \gets Neg^t[n] \bigcup Img_{Neg}^t[n][i]$ 
        \Else \Comment{\textit{Otherwise, add to the \textbf{negatives} of class $n$, per-image \textbf{buffer negatives}}}
            \State $Neg^t[n] \gets Neg^t[n] \bigcup Img_{Buf}^t[i]$
        \EndIf
    \EndFor
\EndFor
\State \textit{Return $Pos^t, Neg^t$} \Comment{\textit{\textbf{Return} positive and negative features at iteration $t$}}
\end{algorithmic} 
\end{algorithm*}

\section*{Acknowledgment}

This research received support by the ERA-NET CHIST-ERA call 2017 project HEAP. It is based upon work supported by the Center for Brains, Minds and Machines (CBMM), funded by NSF STC award CCF-1231216. L. R. acknowledges the financial support of the European Research Council (grant SLING 819789), the AFOSR projects FA9550-18-1-7009, FA9550-17-1-0390 and BAA-AFRL-AFOSR-2016-0007 (European Office of Aerospace Research and Development), and the EU H2020-MSCA-RISE project NoMADS - DLV-777826.

\ifCLASSOPTIONcaptionsoff
  \newpage
\fi



%



\bibliographystyle{unsrt}
\bibliography{bibliography.bib}  

\begin{thebibliography}{10}

\bibitem{ceola2020segm}
Federico Ceola, Elisa Maiettini, Giulia Pasquale, Lorenzo Rosasco, and Lorenzo
  Natale.
\newblock Fast object segmentation learning with kernel-based methods for
  robotics.
\newblock In {\em 2021 IEEE International Conference on Robotics and Automation
  (ICRA)}, pages 13581--13588, 2021.

\bibitem{ceola2020rpn}
Federico Ceola, Elisa Maiettini, Giulia Pasquale, Lorenzo Rosasco, and Lorenzo
  Natale.
\newblock Fast region proposal learning for object detection for robotics.
\newblock {\em arXiv preprint arXiv:2011.12790}, 2020.

\bibitem{xiang2018posecnn}
Yu~Xiang, Tanner Schmidt, Venkatraman Narayanan, and Dieter Fox.
\newblock {PoseCNN}: A convolutional neural network for 6d object pose
  estimation in cluttered scenes.
\newblock In {\em Robotics: Science and Systems (RSS)}, 2018.

\bibitem{hampali2020honnotate}
Shreyas Hampali, Mahdi Rad, Markus Oberweger, and Vincent Lepetit.
\newblock {HOnnotate}: A method for {3D} annotation of hand and object poses.
\newblock In {\em Proceedings of the IEEE/CVF Conference on Computer Vision and
  Pattern Recognition}, pages 3196--3206, 2020.

\bibitem{icub}
Giorgio Metta, Lorenzo Natale, Francesco Nori, Giulio Sandini, David Vernon,
  Luciano Fadiga, Claes von Hofsten, Kerstin Rosander, Manuel Lopes, Jos{\'{e}}
  Santos-Victor, Alexandre Bernardino, and Luis Montesano.
\newblock The {iCub} humanoid robot: an open-systems platform for research in
  cognitive development.
\newblock {\em Neural networks : the official journal of the International
  Neural Network Society}, 23(8-9):1125--34, 1 2010.

\bibitem{ren2015_faster}
Shaoqing Ren, Kaiming He, Ross Girshick, and Jian Sun.
\newblock Faster {R-CNN}: Towards real-time object detection with region
  proposal networks.
\newblock In {\em Neural Information Processing Systems ({NIPS})}, 2015.

\bibitem{dai2016}
Jifeng Dai, Yi~Li, Kaiming He, and Jian Sun.
\newblock {R-FCN}: Object detection via region-based fully convolutional
  networks.
\newblock In D.~D. Lee, M.~Sugiyama, U.~V. Luxburg, I.~Guyon, and R.~Garnett,
  editors, {\em Advances in Neural Information Processing Systems 29}, pages
  379--387. Curran Associates, Inc., 2016.

\bibitem{tan2020efficientdet}
Mingxing Tan, Ruoming Pang, and Quoc~V Le.
\newblock {EfficientDet}: Scalable and efficient object detection.
\newblock In {\em Proceedings of the IEEE/CVF conference on computer vision and
  pattern recognition}, pages 10781--10790, 2020.

\bibitem{Redmon2018}
Joseph Redmon and Ali Farhadi.
\newblock {YOLOv3}: An incremental improvement.
\newblock {\em CoRR}, abs/1804.02767, 2018.

\bibitem{He2017}
Kaiming He, Georgia Gkioxari, Piotr Doll{\'a}r, and Ross~B. Girshick.
\newblock Mask {R-CNN}.
\newblock {\em 2017 IEEE International Conference on Computer Vision (ICCV)},
  pages 2980--2988, 2017.

\bibitem{huang2019mask}
Zhaojin Huang, Lichao Huang, Yongchao Gong, Chang Huang, and Xinggang Wang.
\newblock Mask {Scoring} {R-CNN}.
\newblock In {\em Proceedings of the IEEE/CVF Conference on Computer Vision and
  Pattern Recognition}, pages 6409--6418, 2019.

\bibitem{liu2018path}
Shu Liu, Lu~Qi, Haifang Qin, Jianping Shi, and Jiaya Jia.
\newblock Path aggregation network for instance segmentation.
\newblock In {\em Proceedings of the IEEE conference on computer vision and
  pattern recognition}, pages 8759--8768, 2018.

\bibitem{bolya2019yolact}
Daniel Bolya, Chong Zhou, Fanyi Xiao, and Yong~Jae Lee.
\newblock {YOLACT}: Real-time instance segmentation.
\newblock In {\em Proceedings of the IEEE/CVF International Conference on
  Computer Vision}, pages 9157--9166, 2019.

\bibitem{chen2020blendmask}
Hao Chen, Kunyang Sun, Zhi Tian, Chunhua Shen, Yongming Huang, and Youliang
  Yan.
\newblock {BlendMask}: Top-down meets bottom-up for instance segmentation.
\newblock In {\em Proceedings of the IEEE/CVF conference on computer vision and
  pattern recognition}, pages 8573--8581, 2020.

\bibitem{Lin2017focal}
Tsung{-}Yi Lin, Priya Goyal, Ross~B. Girshick, Kaiming He, and Piotr
  Doll{\'{a}}r.
\newblock Focal loss for dense object detection.
\newblock In {\em {IEEE} International Conference on Computer Vision, {ICCV}
  2017, Venice, Italy, October 22-29, 2017}, pages 2999--3007, 2017.

\bibitem{tian2019fcos}
Zhi Tian, Chunhua Shen, Hao Chen, and Tong He.
\newblock {FCOS}: Fully convolutional one-stage object detection.
\newblock In {\em Proceedings of the IEEE/CVF International Conference on
  Computer Vision}, pages 9627--9636, 2019.

\bibitem{zhou2019objects}
Xingyi Zhou, Dequan Wang, and Philipp Kr{\"a}henb{\"u}hl.
\newblock Objects as points.
\newblock {\em arXiv preprint arXiv:1904.07850}, 2019.

\bibitem{peng2020}
Sida Peng, Wen Jiang, Huaijin Pi, Xiuli Li, Hujun Bao, and Xiaowei Zhou.
\newblock Deep {Snake} for real-time instance segmentation.
\newblock In {\em Proceedings of the IEEE/CVF Conference on Computer Vision and
  Pattern Recognition}, pages 8533--8542, 2020.

\bibitem{gao2019ssap}
Naiyu Gao, Yanhu Shan, Yupei Wang, Xin Zhao, Yinan Yu, Ming Yang, and Kaiqi
  Huang.
\newblock {SSAP}: Single-shot instance segmentation with affinity pyramid.
\newblock In {\em Proceedings of the IEEE/CVF International Conference on
  Computer Vision}, pages 642--651, 2019.

\bibitem{kirillov2017instancecut}
Alexander Kirillov, Evgeny Levinkov, Bjoern Andres, Bogdan Savchynskyy, and
  Carsten Rother.
\newblock {InstanceCut}: from edges to instances with multicut.
\newblock In {\em Proceedings of the IEEE Conference on Computer Vision and
  Pattern Recognition}, pages 5008--5017, 2017.

\bibitem{bai2017}
Min Bai and Raquel Urtasun.
\newblock Deep watershed transform for instance segmentation.
\newblock In {\em Proceedings of the IEEE Conference on Computer Vision and
  Pattern Recognition}, pages 5221--5229, 2017.

\bibitem{pinheiro2016learning}
Pedro~O Pinheiro, Tsung-Yi Lin, Ronan Collobert, and Piotr Doll{\'a}r.
\newblock Learning to refine object segments.
\newblock In {\em European conference on computer vision}, pages 75--91.
  Springer, 2016.

\bibitem{dai2016instance}
Jifeng Dai, Kaiming He, Yi~Li, Shaoqing Ren, and Jian Sun.
\newblock Instance-sensitive fully convolutional networks.
\newblock In {\em European Conference on Computer Vision}, pages 534--549.
  Springer, 2016.

\bibitem{chen2019tensormask}
Xinlei Chen, Ross Girshick, Kaiming He, and Piotr Doll{\'a}r.
\newblock {TensorMask}: A foundation for dense object segmentation.
\newblock In {\em Proceedings of the IEEE/CVF International Conference on
  Computer Vision}, pages 2061--2069, 2019.

\bibitem{8506350}
Xin Shu, Chang Liu, Tong Li, Chunkai Wang, and Cheng Chi.
\newblock A self-supervised learning manipulator grasping approach based on
  instance segmentation.
\newblock {\em IEEE Access}, 6:65055--65064, 2018.

\bibitem{wada2019joint}
Kentaro Wada, Kei Okada, and Masayuki Inaba.
\newblock Joint learning of instance and semantic segmentation for robotic
  pick-and-place with heavy occlusions in clutter.
\newblock In {\em 2019 International Conference on Robotics and Automation
  (ICRA)}, pages 9558--9564. IEEE, 2019.

\bibitem{li2020one}
Andrew Li, Michael Danielczuk, and Ken Goldberg.
\newblock One-shot shape-based amodal-to-modal instance segmentation.
\newblock In {\em 2020 IEEE 16th International Conference on Automation Science
  and Engineering (CASE)}, pages 1375--1382. IEEE, 2020.

\bibitem{li2020learning}
Siyi Li, Jiaji Zhou, Zhenzhong Jia, Dit-Yan Yeung, and Matthew~T. Mason.
\newblock Learning accurate objectness instance segmentation from
  photorealistic rendering for robotic manipulation.
\newblock In Jing Xiao, Torsten Kr{\"o}ger, and Oussama Khatib, editors, {\em
  Proceedings of the 2018 International Symposium on Experimental Robotics},
  pages 245--255, Cham, 2020. Springer International Publishing.

\bibitem{danielczuk2019segmenting}
Michael Danielczuk, Matthew Matl, Saurabh Gupta, Andrew Li, Andrew Lee, Jeffrey
  Mahler, and Ken Goldberg.
\newblock Segmenting unknown 3d objects from real depth images using {Mask
  R-CNN} trained on synthetic data.
\newblock In {\em 2019 International Conference on Robotics and Automation
  (ICRA)}, pages 7283--7290. IEEE, 2019.

\bibitem{xie2020best}
Christopher Xie, Yu~Xiang, Arsalan Mousavian, and Dieter Fox.
\newblock The best of both modes: Separately leveraging rgb and depth for
  unseen object instance segmentation.
\newblock In {\em Conference on robot learning}, pages 1369--1378. PMLR, 2020.

\bibitem{xie2021unseen}
Christopher Xie, Yu~Xiang, Arsalan Mousavian, and Dieter Fox.
\newblock Unseen object instance segmentation for robotic environments.
\newblock {\em IEEE Transactions on Robotics}, 2021.

\bibitem{kuo2019shapemask}
Weicheng Kuo, Anelia Angelova, Jitendra Malik, and Tsung-Yi Lin.
\newblock {ShapeMask}: Learning to segment novel objects by refining shape
  priors.
\newblock In {\em Proceedings of the IEEE/CVF International Conference on
  Computer Vision}, pages 9207--9216, 2019.

\bibitem{pathak2018learning}
Deepak Pathak, Yide Shentu, Dian Chen, Pulkit Agrawal, Trevor Darrell, Sergey
  Levine, and Jitendra Malik.
\newblock Learning instance segmentation by interaction.
\newblock In {\em Proceedings of the IEEE Conference on Computer Vision and
  Pattern Recognition Workshops}, pages 2042--2045, 2018.

\bibitem{eitel2019self}
Andreas Eitel, Nico Hauff, and Wolfram Burgard.
\newblock Self-supervised transfer learning for instance segmentation through
  physical interaction.
\newblock In {\em 2019 IEEE/RSJ International Conference on Intelligent Robots
  and Systems (IROS)}, pages 4020--4026. IEEE, 2019.

\bibitem{DAVIS2020-Semi-Supervised-1st}
B.~Zhang P.~Zhang, L.~Hu and P.~Pan.
\newblock Spatial constrained memory network for semi-supervised video object
  segmentation.
\newblock {\em The 2020 DAVIS Challenge on Video Object Segmentation - CVPR
  Workshops}, 2020.

\bibitem{DAVIS2020-Unsupervised-1st}
V.~Goel S.~Garg and S.~Kumar.
\newblock Unsupervised video object segmentation using online mask selection
  and space-time memory networks.
\newblock {\em The 2020 DAVIS Challenge on Video Object Segmentation - CVPR
  Workshops}, 2020.

\bibitem{voigtlaender2017online}
Paul Voigtlaender and Bastian Leibe.
\newblock Online adaptation of convolutional neural networks for video object
  segmentation.
\newblock {\em arXiv preprint arXiv:1706.09364}, 2017.

\bibitem{siam2019video}
Mennatullah Siam, Chen Jiang, Steven Lu, Laura Petrich, Mahmoud Gamal, Mohamed
  Elhoseiny, and Martin Jagersand.
\newblock Video object segmentation using teacher-student adaptation in a human
  robot interaction (hri) setting.
\newblock In {\em 2019 International Conference on Robotics and Automation
  (ICRA)}, pages 50--56. IEEE, 2019.

\bibitem{7989364}
Raffaello Camoriano, Giulia Pasquale, Carlo Ciliberto, Lorenzo Natale, Lorenzo
  Rosasco, and Giorgio Metta.
\newblock Incremental robot learning of new objects with fixed update time.
\newblock In {\em 2017 IEEE International Conference on Robotics and Automation
  (ICRA)}, pages 3207--3214, 2017.

\bibitem{MALTONI201956}
Davide Maltoni and Vincenzo Lomonaco.
\newblock Continuous learning in single-incremental-task scenarios.
\newblock {\em Neural Networks}, 116:56--73, 2019.

\bibitem{shmelkov2017incremental}
Konstantin Shmelkov, Cordelia Schmid, and Karteek Alahari.
\newblock Incremental learning of object detectors without catastrophic
  forgetting.
\newblock In {\em Proceedings of the IEEE international conference on computer
  vision}, pages 3400--3409, 2017.

\bibitem{perez2020incremental}
Juan-Manuel Perez-Rua, Xiatian Zhu, Timothy~M Hospedales, and Tao Xiang.
\newblock Incremental few-shot object detection.
\newblock In {\em Proceedings of the IEEE/CVF Conference on Computer Vision and
  Pattern Recognition}, pages 13846--13855, 2020.

\bibitem{Michieli_2019_ICCV}
Umberto Michieli and Pietro Zanuttigh.
\newblock Incremental learning techniques for semantic segmentation.
\newblock In {\em Proceedings of the IEEE/CVF International Conference on
  Computer Vision (ICCV) Workshops}, Oct 2019.

\bibitem{falkon2018}
Alessandro Rudi, Luigi Carratino, and Lorenzo Rosasco.
\newblock {FALKON}: An optimal large scale kernel method.
\newblock In I.~Guyon, U.~V. Luxburg, S.~Bengio, H.~Wallach, R.~Fergus,
  S.~Vishwanathan, and R.~Garnett, editors, {\em Advances in Neural Information
  Processing Systems 30}, pages 3888--3898. Curran Associates, Inc., 2017.

\bibitem{falkonlibrary2020}
Giacomo Meanti, Luigi Carratino, Lorenzo Rosasco, and Alessandro Rudi.
\newblock Kernel methods through the roof: handling billions of points
  efficiently.
\newblock {\em arXiv preprint arXiv:2006.10350v1}, 2020.

\bibitem{girshick2014_rcnn}
Ross Girshick, Jeff Donahue, Trevor Darrell, and Jitendra Malik.
\newblock Rich feature hierarchies for accurate object detection and semantic
  segmentation.
\newblock In {\em Proceedings of the IEEE Conference on Computer Vision and
  Pattern Recognition ({CVPR})}, 2014.

\bibitem{maiettini2018}
E.~Maiettini, G.~Pasquale, L.~Rosasco, and L.~Natale.
\newblock Speeding-up object detection training for robotics with {FALKON}.
\newblock In {\em 2018 IEEE/RSJ International Conference on Intelligent Robots
  and Systems (IROS)}, Oct 2018.

\bibitem{maiettini2019a}
Elisa Maiettini, Giulia Pasquale, Lorenzo Rosasco, and Lorenzo Natale.
\newblock On-line object detection: a robotics challenge.
\newblock {\em Autonomous Robots}, Nov 2019.

\bibitem{Sung1996}
Kah~Kay Sung.
\newblock {\em Learning and Example Selection for Object and Pattern
  Detection}.
\newblock PhD thesis, Massachusetts Institute of Technology, Cambridge, MA,
  USA, 1996.
\newblock AAI0800657.

\bibitem{Lin2017_fpn}
Tsung-Yi Lin, Piotr Doll{\'a}r, Ross Girshick, Kaiming He, Bharath Hariharan,
  and Serge Belongie.
\newblock Feature pyramid networks for object detection.
\newblock In {\em Proceedings of the IEEE conference on computer vision and
  pattern recognition}, pages 2117--2125, 2017.

\bibitem{He2015}
Kaiming He, Xiangyu Zhang, Shaoqing Ren, and Jian Sun.
\newblock Deep residual learning for image recognition.
\newblock In {\em Proceedings of the IEEE conference on computer vision and
  pattern recognition}, pages 770--778, 2016.

\bibitem{pascal2010}
M.~Everingham, L.~Van~Gool, C.~K.~I. Williams, J.~Winn, and A.~Zisserman.
\newblock The {Pascal} visual object classes {(VOC)} challenge.
\newblock {\em International Journal of Computer Vision}, 88(2):303--338, June
  2010.

\bibitem{coco}
Tsung-Yi Lin, Michael Maire, Serge Belongie, James Hays, Pietro Perona, Deva
  Ramanan, Piotr Dollár, and C.~Lawrence Zitnick.
\newblock Microsoft {COCO}: Common objects in context.
\newblock In {\em European Conference on Computer Vision (ECCV)}, Zürich,
  2014.
\newblock Oral.

\bibitem{calli2015ycb}
Berk Calli, Arjun Singh, Aaron Walsman, Siddhartha Srinivasa, Pieter Abbeel,
  and Aaron~M Dollar.
\newblock The {YCB} object and model set: Towards common benchmarks for
  manipulation research.
\newblock In {\em 2015 international conference on advanced robotics (ICAR)},
  pages 510--517. IEEE, 2015.

\bibitem{Mettayarp}
G.~Metta, P.~Fitzpatrick, and L.~Natale.
\newblock {YARP}: Yet another robot platform.
\newblock {\em International Journal of Advanced Robotics Systems},
  3(1):43--48, 2006.

\bibitem{10.3389/frobt.2016.00035}
Giulia Pasquale, Tanis Mar, Carlo Ciliberto, Lorenzo Rosasco, and Lorenzo
  Natale.
\newblock Enabling depth-driven visual attention on the {iCub} humanoid robot:
  Instructions for use and new perspectives.
\newblock {\em Frontiers in Robotics and AI}, 3:35, 2016.

\end{thebibliography}

%


\begin{IEEEbiography}[{\includegraphics[width=1in,height=1.25in,clip,keepaspectratio]{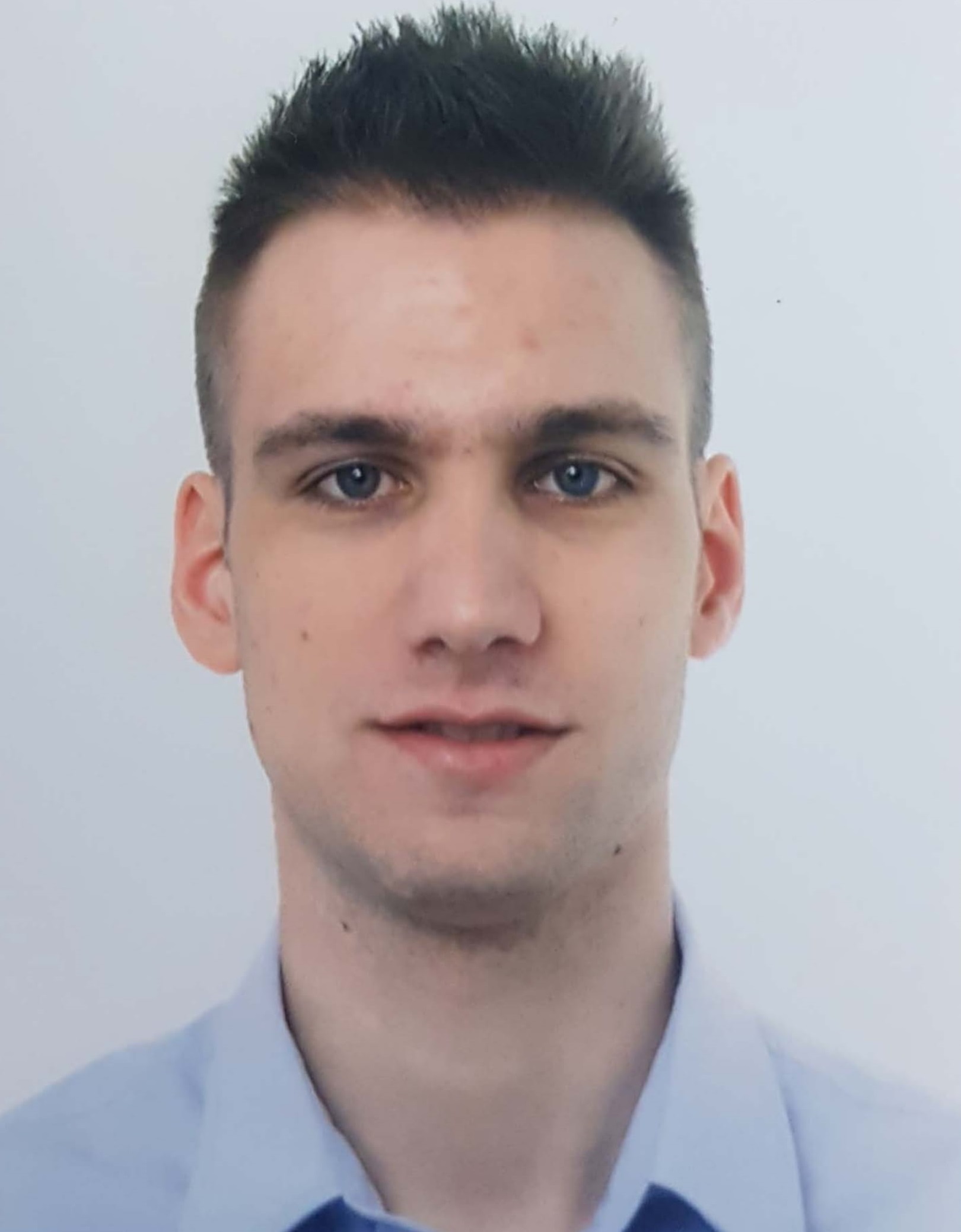}}]{Federico Ceola}
is a Ph.D. student at the Humanoid Sensing and Perception (HSP) group at the Istituto Italiano di Tecnologia and at the Laboratory for Computational and Statistical Learning (LCSL) at the University of Genova. He received his Bachelor's Degree in Information Engineering and his Master's Degree (with honors) in Computer Engineering at the University of Padova in 2016 and in 2019, respectively. His research interests lie in the intersection of Robotics, Computer Vision and Machine Learning.
\end{IEEEbiography}

\begin{IEEEbiography}[{\includegraphics[width=1in,height=1.25in,clip,keepaspectratio]{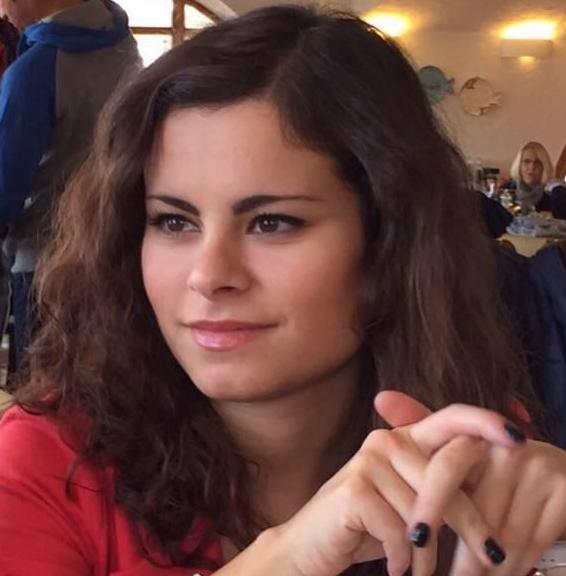}}]{Elisa Maiettini}
is a Post Doctoral researcher in the Humanoid Sensing and Perception (HSP) research line. She graduated in Software and Electronics Engineering at the University of Perugia, Italy, in 2013 and she obtained an M.D. with honors in Software and Automation Engineering at the same university in 2016. She received the Ph. D. in Bioengineering and Robotics from 2016 to 2020, at the Istituto Italiano di Tecnologia. The main fields of her research are: computer vision, machine learning and humanoid robotics.
\end{IEEEbiography}

\begin{IEEEbiography}[{\includegraphics[width=1in,height=1.25in,clip,keepaspectratio]{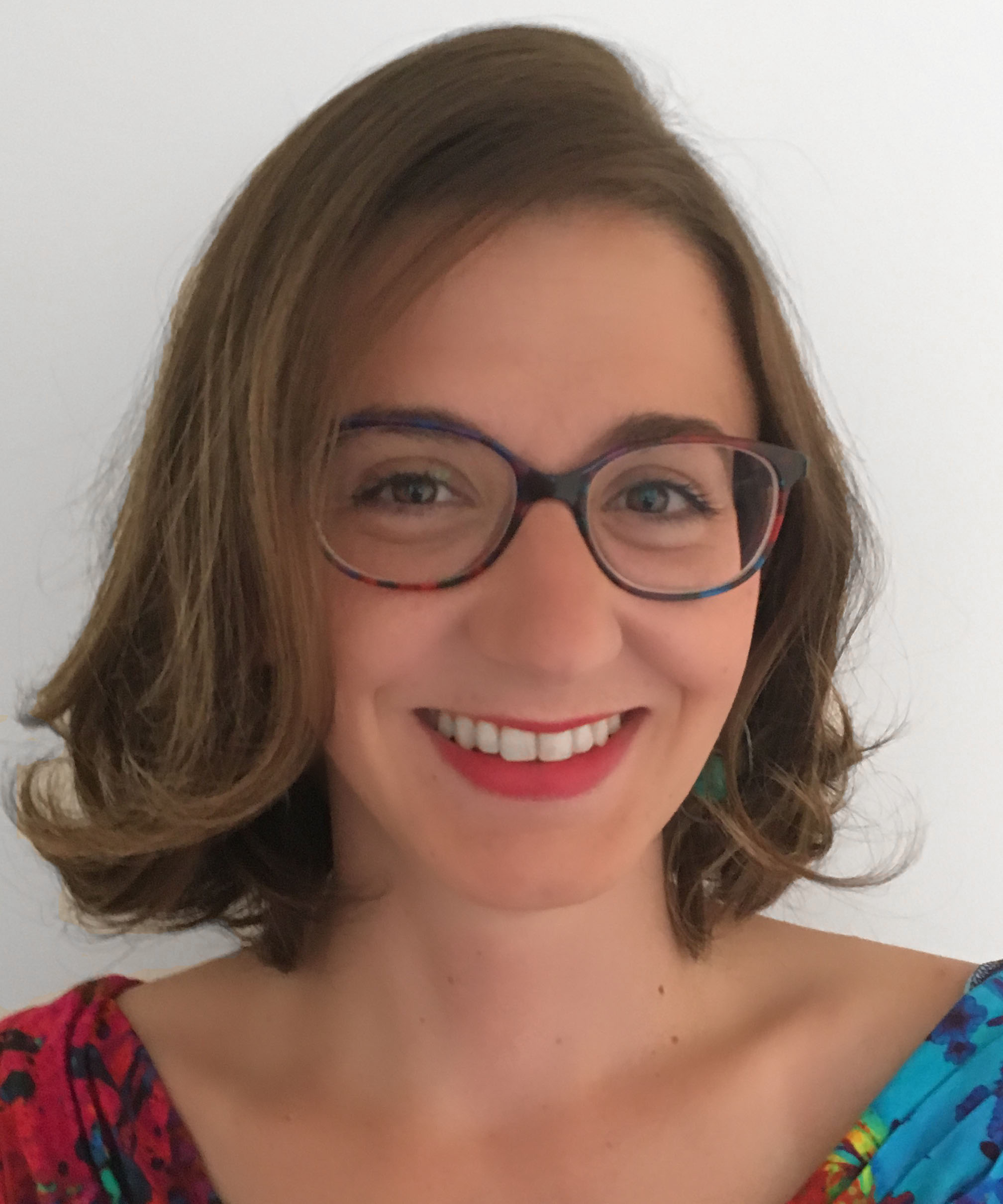}}]{Giulia Pasquale}
is Senior Technician at the Istituto Italiano di Tecnologia, Humanoid Sensing and Perception (IIT HSP) research line. She received her B.Sc. degree in Biomedical Engineering (with honors) and M.Sc. Degree in Bioengineering (with honors), in 2010 and 2013, at the University of Genoa. In 2013 she was research fellow at the National Research Council of Italy. She then obtained a Ph.D. in Bioengineering and Robotics within a collaboration between IIT and the University of Genoa from 2014 to 2017. She was later Postdoctoral Researcher at IIT HSP until 2019. Her research focuses on the development of visual object recognition and localization systems for robotic platforms, standing at the intersection between machine/deep learning, computer vision and robotics.\end{IEEEbiography}

\begin{IEEEbiography}[{\includegraphics[width=1in,height=1.25in,clip,keepaspectratio]{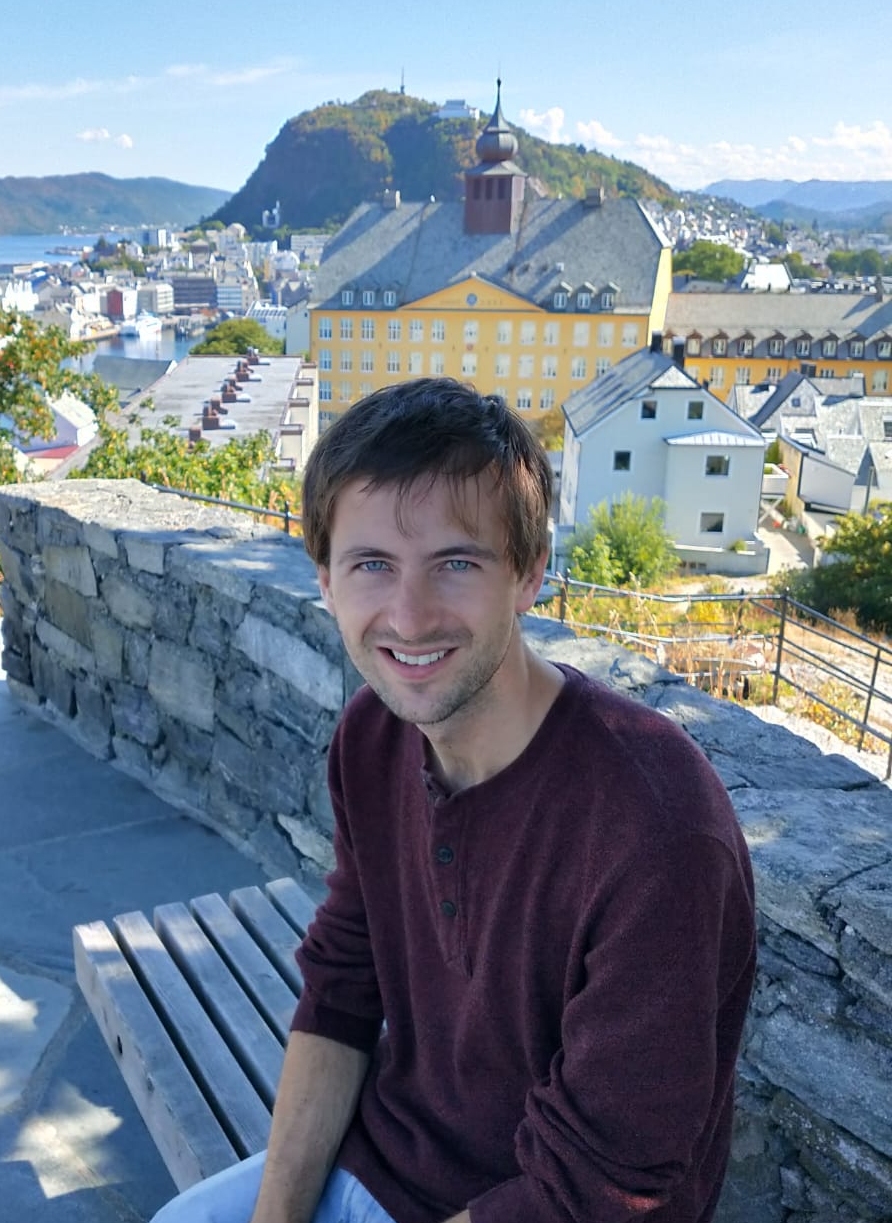}}]{Giacomo Meanti}
is a second-year Ph.D. student in Computer Science. After his BSc at the University of Southampton, he received a MSc degree from ETH Zurich in 2018. He worked as a modelling expert at a startup in Switzerland, and is now working on his PhD at the University of Genova under the supervision of Lorenzo Rosasco.
\end{IEEEbiography}

\begin{IEEEbiography}[{\includegraphics[width=1in,height=1.25in,clip,keepaspectratio]{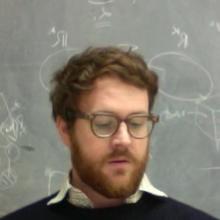}}]{Lorenzo Rosasco}
is full professor at the University of Genova, Italy. He is also affiliated with the Massachusetts Institute of Technology (MIT), where is a visiting professor, and with the Istituto Italiano di Tecnologia (IIT), where he is an external collaborator. He is leading the efforts to establish the Laboratory for Computational and Statistical Learning (LCSL), born from a collaborative agreement between IIT and MIT. He received his PhD from the University of Genova in 2006. Between 2006 and 2009 he was a postdoctoral fellow at Center for Biological and Computational Learning (CBCL) at MIT working with Tomaso Poggio. His research focuses on studying theory and algorithms for machine learning. He has developed and analyzed methods to learn from small as well as large samples of high dimensional data, using analytical and probabilistic tools, within a multidisciplinary approach drawing concepts and techniques primarily from computer science but also from statistics, engineering and applied mathematics.
\end{IEEEbiography}

\begin{IEEEbiography}[{\includegraphics[width=1in,height=1.25in,clip,keepaspectratio]{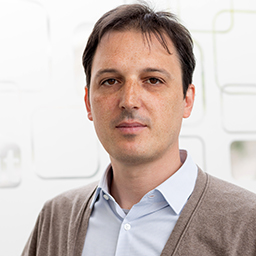}}]{Lorenzo Natale}
has a degree in Electronic Engineering and Ph.D. in Robotics. He was later postdoctoral researcher at the MIT Computer Science and Artificial Intelligence Laboratory. At the moment he is seniore researcher in the Istituto Italiano di Tecnologia, Genova, Italy, where he leads the Humanoid Sensing and Perception group. His research interests include artificial perception and software architectures for robotics. Dr. Natale served as the Program Chair of ICDL-Epirob 2014 and HAI 2017. He is Specialty Chief Editor for the Humanoid Robotics Section of Frontiers in Robotics and AI, associate editor for IEEE-Transactions on Robotics. He is also Ellis Fellow and Core Faculty of the Ellis Genoa Unit.
\end{IEEEbiography}







\end{document}